\documentclass[final]{cvpr}

\usepackage{times}
\usepackage{epsfig}
\usepackage{graphicx}
\usepackage{amsmath}
\usepackage{amssymb}
\usepackage{textcomp}
\usepackage{subcaption}
\usepackage{algorithm}
\usepackage{booktabs}
\usepackage[noend]{algpseudocode}
\usepackage{tabularx}
\usepackage{tablefootnote}
\usepackage{listings}
\usepackage{multirow}

\usepackage[pagebackref=true,breaklinks=true,colorlinks,bookmarks=false]{hyperref}

\def\cvprPaperID{1874} % *** Enter the CVPR Paper ID here
\def\confYear{CVPR 2021}

\begin{document}

%%%%%%%%% TITLE
\title{FrostNet: Towards Quantization-Aware Network Architecture Search} 

\author{Taehoon~Kim\\
Machine Learning Laboratory\\
Sogang University\\
{\tt\small taehoonkim@sogang.ac.kr}
\and
Youngjoon~Yoo\\
 Clova AI Research,\\
NAVER Corporation\\
{\tt\small youngjoon.yoo@navercorp.com}

\and
Jihoon~Yang\\
 Machine Learning Laboratory\\
 Sogang University\\
{\tt\small yangjh@sogang.ac.kr}
}

\maketitle
%\thispagestyle{empty}

%%%%%%%%% ABSTRACT
\begin{abstract}
\texttt{INT8} quantization has become one of the standard techniques for deploying convolutional neural networks (CNNs) on edge devices to reduce the memory and computational resource usages. 
By analyzing quantized performances of existing mobile-target network architectures, we can raise an issue regarding the importance of network architecture for optimal \texttt{INT8} quantization. 
In this paper, we present a new network architecture search (NAS) procedure to find a network that guarantees both full-precision (\texttt{FLOAT32}) and quantized (\texttt{INT8}) performances. 
We first propose critical but straightforward optimization method which enables quantization-aware training (QAT) : floating-point statistic assisting~(\textit{StatAssist}) and stochastic gradient boosting (\textit{GradBoost}). 
By integrating the gradient-based NAS with \textit{StatAssist} and \textit{GradBoost}, we discovered  a quantization-efficient network building block, \textit{Frost} bottleneck. 
Furthermore, we used \textit{Frost} bottleneck as the building block for hardware-aware NAS to obtain quantization-efficient networks, \textit{FrostNets}, which show improved quantization performances compared to other mobile-target networks while maintaining competitive \texttt{FLOAT32} performance. 
Our \textit{FrostNets} achieve higher recognition accuracy than existing CNNs with comparable latency when quantized, due to higher latency reduction rate (average 65\%).
\end{abstract}

\section{Introduction}
\label{sec:introduction}
\begin{figure}[!tb]
  \centering
\includegraphics[width=0.95\linewidth]{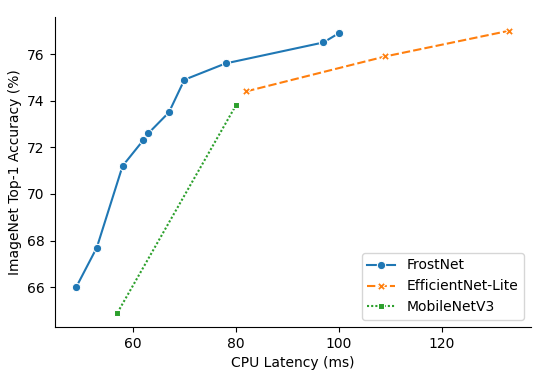}
\caption{
CPU Latency vs. ImageNet Top-1 Accuracy in \texttt{INT8} quantized setting.
All latencies were measured on the same device with each model's original input resolution. Our FrostNets show outperforming trade-offs between latency and accuracy compared to other lightweight models.}% Best viewed in color.}
\label{fig:acc_latency}
\vspace{-4mm}
\end{figure}
\begin{table}[!tb]
\addtolength{\tabcolsep}{-2pt}
\centering
  \begin{tabular}{lcccc}
    \toprule
Model          & CPU(f) & CPU(q) & Rate \\ 
\midrule
%MobileNetV2~\cite{mobilenetv2}                 & 190 ms & 57 ms & $70.0\%$&\\
MobileNetV3-Large~\cite{howard2019searching}   & 197 ms & 80 ms & $59.3\%$&\\
MobileNetV3-Small~\cite{howard2019searching}  & 113 ms & 57 ms & $49.5\%$&\\
EfficientNet-Lite0~\cite{Tan2019EfficientNetRM}  & 197 ms & 82 ms & $58.4\%$&\\
EfficientNet-Lite1~\cite{Tan2019EfficientNetRM}  & 273 ms & 109 ms & $60.1\%$&\\
EfficientNet-Lite2~\cite{Tan2019EfficientNetRM}  & 350 ms & 133 ms & $62.0\%$&\\
FrostNet-Large                                & 234 ms & 78 ms & $\textbf{66.7\%}$&\\
FrostNet-Base                                 & 233 ms & 70 ms & $\textbf{69.9\%}$&\\
FrostNet-Small                                & 185 ms & 62 ms & $\textbf{66.5\%}$&\\
\midrule
ShuffleNetV2~\cite{zhang2017shufflenet} & 153 ms & 237 ms & $-54.9\%$&\\
GhostNet~\cite{han2020ghostnet} & 229 ms & 381 ms & $-66.3\%$&\\
\midrule
\end{tabular}

\caption
{
Inference latency results of different lightweight models in \texttt{FLOAT32} full-precision and \texttt{INT8} quantized settings. \textbf{CPU(f)}: CPU latency in \texttt{FLOAT32}. 
\textbf{CPU(q)}: CPU latency in \texttt{INT8}. 
\textbf{Rate}: Latency reduction rate from \texttt{FLOAT32} to \texttt{INT8}. All latencies were measured on the same device with each model's original input resolution. Our FrostNets show better latency reduction rates compared to other lightweight models.
} 
\label{table:quant_latency}
\end{table}
Quantization of the weight and activation of the deep model has been a promising approach to reduce the model complexity, along with other techniques such as pruning~\cite{lecun1990optimal} and distillation~\cite{hinton2015distilling}. Previous studies, both on weight-only quantization and weight-activation quantization, have achieved meaningful progress on computer vision tasks.
Especially, the scalar~(\texttt{INT8}) quantization provides practically applicable performances with enhanced latency, and has become a new standard technique for the deployment of mobile-target lightweight networks on edge devices.  However, the actual amount of performance gain that can be obtained from quantization varies greatly depending on the network architecture. From the results in Table ~\ref{table:quant_latency}, we can raise an issue regarding effective quantization of current lightweight network architectures: \textit{What is quantization-aware architecture design scheme?} 

Previous methods~\cite{fan2020training,Li2019quantizedObjectDet,jacob2018quantization,howard2019searching} report meaningful latency-enhancement of the convolutional block, but this couldn't always lead to the overall speed-up. While the quantization statistics of weight layers are fixed, the quantization statistics of normalization and activation layers constantly changes according to the input. We cannot expect significant latency improvement if we were to update these numbers for each input mini-batch. The integration process of the weight, normalization, and activation into a single layer, called \textit{layer fusion} \cite{jacob2018quantization}, is essential to boosting quantization performances. Since the \textit{fusible} combinations of operations are very limited, the layer fusion restricts the selection of the normalization and activation for network components. For example, Tan \etal~\cite{Tan2019EfficientNetRM} removed squeeze-and-excite (SE) \cite{Hu2018SqueezeandExcitationN} and replaced all \textit{swish} \cite{ramachandran2017searching} with \textit{ReLU6} \cite{mobilenetv2} for their mobile/IoT friendly EfficientNet-Lite.

This paper describes the series of approaches we took to find the quantization-friendly network architecture, \textit{FrostNet}. 
As shown in Figure~\ref{fig:acc_latency}, our FrostNets guarantee outperforming trade-offs between accuracy and latency in the \texttt{INT8} quantized setting. We first found some prototypes for a new quantization-friendly network building block by running gradient-based network architecture search (NAS) \cite{liu2018darts,chen2019progressive, xu2020pcdarts}, in the \textit{quantization-aware-training} (QAT)~\cite{jacob2018quantization} setting. Since current QAT mechanisms doesn't support the \textit{fake-quantized} training from scratch \cite{jacob2018quantization,DBLP:journals/corr/abs-1806-08342}, we additionally propose intensive and flexible strategies that enable the scratch QAT. After manual pruning process described in Section~\ref{sec:pruning_process}, we ran hardware-aware NAS \cite{Tan2019MnasNetPN,cai2018proxylessnas} again in the scratch QAT setting to find FrostNet architectures specified in Table~\ref{table:frost_arch}.

Specifically, our contributions for the quantization-aware network architecture search are summarized as follows:
\begin{itemize}
    \item We introduce floating-point statistic assisting~(\textit{StatAssist}) and stochastic gradient boosting (\textit{GradBoost}) for stable and cost-saving QAT from scratch. Our method leads to successful QAT in various tasks including classification~\cite{he2016deep,mobilenetv2,Ma_2018_ECCV}, object detection~\cite{mobilenetv2,tdsod}, segmentation~\cite{mehta2018espnet,mehta2018espnetv2,mobilenetv2,howard2019searching}, and style transfer~\cite{pix2pix}.
    \item By combining StatAssist and GradBoost QAT with various NAS algorithms, we developed a novel network architecture, \textit{FrostNet}, which shows improved accuracy-latency trade-offs compared to other state-of-the-art lightweight networks while maintaining competitive full-precision performances.
    \item Our FrostNet also highly surpass other mobile-target networks when used as a drop-in replacement for the backbone feature extractor, achieving 33.7 mAP (FrostNet-Small-Retinanet) in object detection task with MS COCO 2017~\cite{lin2015microsoft} \textit{val} split. 
\end{itemize}
 Section \ref{sec:background} briefly reviews mobile-target network building blocks, network architecture search algorithms, and network quantization from previous works. Section \ref{sec:statassist_gradboost} describes the StatAssist and GradBoost algorithms for QAT from scratch without quantized performance degradation. Section \ref{sec:FrostNet} explains how we designed a new building block to improve its quantization efficiency and introduces the final FrostNet architecture. Section \ref{sec:experiments} shows extensive experiments on different computer vision tasks to demonstrate main contributions of this paper.  Section \ref{sec:conclusion} summarizes our paper with a meaningful conclusion.

\section{Background}
\label{sec:background}
To run deep neural networks on mobile devices, designing deep neural network architecture for the optimal trade-off between accuracy and efficiency has been an active research area in recent years. Both handcrafted structures and automatically found structures with NAS have played important roles in providing on-device deep learning experiences with reduced latency and power consumption. 

\subsection{Efficient Mobile Building Blocks}
Mobile-target models have been built on computationally efficient building blocks. Depth-wise separable convolution was first introduced in MobileNetV1 \cite{howard2017mobilenets} to replace traditional convolution layers. Depth-wise separable convolutions effectively factorized traditional convolution operation with the competitive functionality by separating spatial filtering from the feature generation mechanism. Separated feature generation are done by heavier $1 \times 1$ point-wise convolutions which is usually placed right after each depth-wise convolution.

MobileNetV2 \cite{mobilenetv2} introduced an improved version of mobile building block with the linear bottleneck and inverted residual structure, making the computation more efficient. This block design is defined by a $1\times1$ expansion convolution followed by a $n\times n$ depth-wise convolution and a $1\times1$ linear projection layer. This structure maintains a compact representation at the input and the output, expanding to a higher-dimensional feature space internally.
%The input and out put are connected with a residual connection. Since this structure maintains a compact representation at the input and the output while expanding to a higher-dimensional feature space internally, recent mobile network architectures use MobilenetV2 block as their basic building block. 

MnasNet \cite{Tan2019MnasNetPN} further enhances the performance of the \textit{inverted residual bottleneck } (\textit{MBConv}) \cite{mobilenetv2} by adding lightweight attention modules based on squeeze and excitation \cite{Hu2018SqueezeandExcitationN} into the original bottleneck structure. MobilenetV3 \cite{howard2019searching} further replace the non-linear activation function of the block with \textit{swish} \cite{ramachandran2017searching} non-linearity for better accuracy. Using \textit{MBConv} equipped with \textit{squeeze-and-excite} (SE) \cite{Hu2018SqueezeandExcitationN} and \textit{swish} \cite{ramachandran2017searching}, EfficientNet \cite{Tan2019EfficientNetRM} achieved state-of-the-art performance in Imagenet \cite{imagenet_cvpr09} classification. However, the \textit{squeeze-and-excite} and \textit{swish} were removed from the mobile/IoT friendly version of EfficientNet (EfficientNet-Lite) to support mobile accelerators and post-quantization. 

\subsection{Network Architecture Search} 
While most of network architectures in early stages of research were manually designed and tested by humans, the automated architecture design process, network architecture search (NAS) \cite{zoph2018learning,pham2018efficient,Tan2019MnasNetPN}, has been reducing man-power spent for trials and errors in recent works. Reinforcement learning (RL) was first introduced as a key component to search efficient architectures with practical accuracy. 

To reduce the computational cost, gradient-based NAS algorithms \cite{liu2018darts,chen2019progressive,xu2020pcdarts} were proposed and have shown comparable results to architectures searched in a fully configurable search space. Focusing on layer-wise search by using existing network building blocks, hardware-aware NAS algorithms \cite{Tan2019MnasNetPN,cai2018proxylessnas} also presents efficient network architectures with great accuracy and efficiency trade-offs.

\subsection{Network Quantization}
Network quantization requires approximating the weight parameters $W\in\mathcal{R}$ and activation $a\in\mathcal{R}$ of the network $F$ to $W_q\in\mathcal{R}_q$ and $a_q\in\mathcal{R}_q$, where the space $\mathcal{R}$ and $\mathcal{R}_q$ each denotes the space represented by \texttt{FLOAT32} and \texttt{INT8} precision.
From Jacob~\etal~\cite{jacob2018quantization}, the process of approximating the original value $x\in\mathcal{R}$ to $x_q\in\mathcal{R}_q$ can be represented as:
\begin{equation} 
\label{eq:quantization}
\begin{aligned}
&x_q = A(x; S(x)), \\
&S(x) = \{min(x), max(x), zero(x)\}
\end{aligned}
\end{equation}
Here, the approximation function $A$ and its inverse function $A^{-1}$ are defined by the same parameters $S(x)$. 
It means that if we store the quantization statistics $S(x)$ including minimum, maximum, and zero point of $x$, we can convert $x$ to $x_q$ and revert $x_q$ to $x$. 
Now we assume the multiplication * of the vector $x_{1q} = A(x_1)$ and $x_{2q} = A(x_2)$, where the both vectors lie on $\mathcal{R}_q$. Then, also from Jacob~\etal~\cite{jacob2018quantization}, the resultant value $v\in\mathcal{R}$ is formulated as,
\begin{equation} 
\label{eq:quantization_vector_mul}
\begin{aligned}
&v = x_1*x_2 \simeq A^{-1}(q(x_{1q} * x_{2q}); S(v)),\\
&S(v) = f(S(x_1), S(x_2))
\end{aligned}
\end{equation}
where the function $f$ derives the statistics $S_v$ from $S(x_1)$ and $S(x_2)$ and the function $q$ only includes lower-bit calculation. 
From the equation, we can replace the \texttt{FLOAT32} operation $x_1*x_2$ to \texttt{INT8} operation. Ideally, this provides faster \texttt{INT8} operation and hence faster calculation.

\paragraph{Static Quantization}
\label{sec:static_quant}
Despite the theoretical speed-enhancement, achieving the enhancement by the network quantization is not straightforward. 
One main reason is the quantization of the activation $a$. At the inference time, we can easily get the quantization statistics of weights $S(W)$ since the value of $W$ is fixed. In contrast, the quantization statistics of the activation $S(a)$ constantly change according to the input value of $F$. Since replacing the floating-point operation $x$ with the lower-bit operation $x_q$ always requires $S(x)$, a special workaround is essential for the activation quantization.

Instead of calculating the quantization statistics dynamically, approximating $S(a)$ with the pre-calculated the statistics from a number of samples $x_i\in{X}, i=1,...,N$ can be one solution to detour the problem, and called the \textit{static quantization}~\cite{jacob2018quantization}. In the static quantization process, the approximation function $A$ uses the  pre-calculated statistics $S_{static} = \{s_{min}, s_{max}, s_{zero}\}$, where the statistic is from the set of samples $X$. Since we fix the statistics, there exists a sample $x_{new}$ such that $x_{new}\notin[s_{min}, s_{max}]$, and we truncate the sample to the bound. The static quantization including the calibration of the quantization statistics are also called \textit{post-quantization}~\cite{jacob2018quantization}. As we briefly mentioned in Section~\ref{sec:introduction}, quantization methods also requires \textit{layer fusion}, integration of the convolution, normalization, and activation, for enhanced quantized latency.

\begin{figure*}[!htb]
  \centering
\begin{subfigure}{0.3\textwidth}
  \centering
\includegraphics[width=\linewidth]{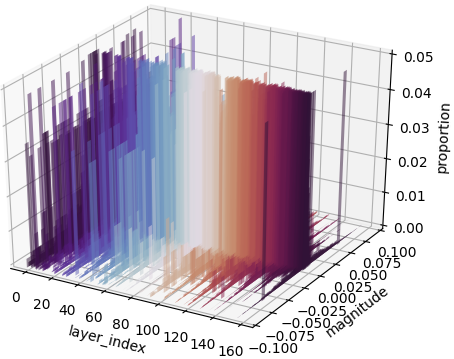}
\caption{FP Training}
\label{fig:1_1}
\end{subfigure}
\begin{subfigure}{0.3\textwidth}
  \centering
\includegraphics[width=\linewidth]{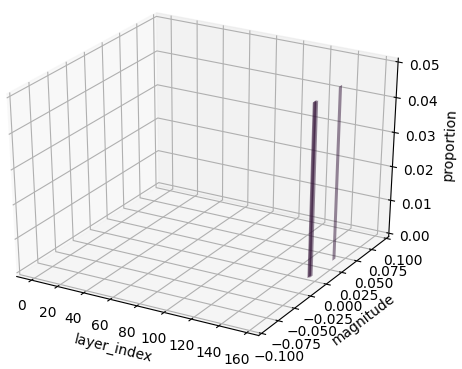}
\caption{Scratch QAT}
\label{fig:1_2}
\end{subfigure}
\begin{subfigure}{0.3\textwidth}
  \centering
\includegraphics[width=\linewidth]{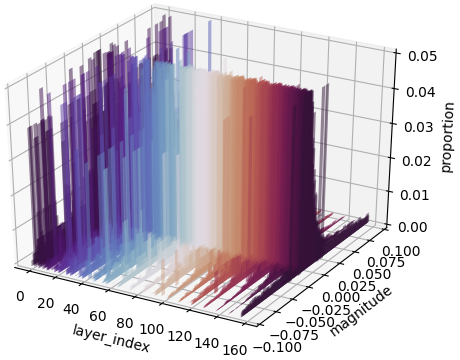}
\caption{StatAssist + GradBoost}
\label{fig:1_3}
\end{subfigure}
\vspace{-2mm}
\caption{Histograms of gradients at the layers of a MobileNetV2~\cite{mobilenetv2} on the CIFAR-10~\cite{cifar} dataset during the $10^{th}$ epoch of the full-precision~(FP) training~(\textit{left}), original scratch quantization-aware training~(QAT)~\cite{jacob2018quantization}~(\textit{middle}), and proposed StatAssist + GradBoost QAT~(\textit{right}). While the scratch QAT model~(\textit{middle}) suffers from the vanishing gradient problem~\cite{grad_vanishing} due to the gradient approximation error 
caused by the straight through estimator~(STE)~\cite{DBLP:journals/corr/BengioLC13}, the StatAssist + GradBoost QAT model~(\textit{right}) shows a distribution of gradients similar to that of the FP-trained counterpart~(\textit{left}), thus provides a comparable performance to the FP baselines.}
\label{fig:1}
\end{figure*}
\paragraph{Quantization-Aware Training}
To alleviate the performance degradation from post-quantization, Jacob \etal propose \textit{quantization-aware training} (QAT)~\cite{jacob2018quantization} %as a method 
to fine-tune the network parameters. In training phase, QAT converts the convolutional block to the fake-quantization module, which mimics the fixed-point computation of the quantized module with the floating-point arithmetic. In the inference phase, each fake-quantized module is actually converted to the lower-bit (\texttt{INT8}) quantized counterpart using the statistics and the weight value of the fake-quantized module.

The optimization of QAT is reported to be unstable~\cite{fan2020training} due to the approximation errors occurred in the fake-quantization with the straight through estimator (STE)~\cite{DBLP:journals/corr/BengioLC13}. This instability restricts the applicability of QAT as a fine-tuning process with pre-trained \texttt{FLOAT32} weights. In Section \ref{sec:statassist_gradboost}, we study the causes of fragility in training by analyzing the gradients and propose a novel method for stable QAT from scratch.

\section{StatAssist and GradBoost}
\label{sec:statassist_gradboost}
In this section, we propose strategies for efficient quantization-aware training from scratch with better quantization performance , as well as reducing training cost. Our proposal tackles two common factors that lead to the failure of QAT from scratch: 1) the gradient approximation error and 2) the gradient instability from the inaccurate initial quantization statistics.

%\subsection{Approximation Error and Gradient Computation}
\subsection{Gradient Approximation Error}
Let the quantity $g(W)$ be the gradient computed for the weight $W$ by the floating point precision. Then, in each update step $t$, the weight $W_t$ is updated as follows:
\begin{equation} 
\label{eq:gradient_update}
W_{t+1} = W_{t} + \eta g(W_{t}) + \beta m_{t}
\end{equation}

\noindent where the term $m_{t}$ denotes the momentum statistics accumulating the traces of the gradient computed in previous time-steps. The term $\eta$ governs the learning rate of the model training. 
In QAT setting, the fake quantization module approximates the process by the function $A(W_t ; S_t)$ in section~\ref{sec:static_quant}, as:
\begin{equation} 
\label{eq:gradient_update_approx}
W_{t+1,q} = W_{t,q} + \eta g(W_{t,q}) + \beta m_{t,q}
\end{equation}
The term $S_t$ denotes the quantization statistics of $W_t$.
The quantization step of $W_t$ includes the value clipping by the $s_{min}$ and $s_{max}$ of the quantization statistics $S_{static}$. This let the calculation $g(W_{t,q})$ occur erroneous approximation, and propagated to the downstream layers invoking gradient vanishing, as in Figure~\ref{fig:1_2}.
%Also, the gradient approximation error and the statistics update form a negative feedback loop amplifying the error. 
The inaccurate calculation of the gradient invokes the inaccurate statistic update, and this inaccurate statistic again induces the inaccurate gradient calculation, forming a feedback loop which amplifies the error.

We suggest that the error amplification can be prevented by assigning a proper momentum value $\hat{m}_t$.
%, as in Figure~\ref{fig:2_3}. 
If the momentum has a proper weight update direction, the weight $W_{t+1}$ of Equation~\ref{eq:gradient_update_approx} will ignore the inaccurate gradient $g(W_{t,q})$.
In this case, we can expect the statistics $S_{static}$ in the next time step $t+1$ to become more accurate than those in $t$. This positive feedback reduces the gradient update error as well as accumulates the statistics from full-precision (\texttt{FLOAT32}). 
In the previous QAT case, the use of the pre-trained weight and the statistic calibration (and freeze) helped reducing the initial gradient computation error. 
Still, the learning rate value $\eta$ is restricted to be small.

\textit{Then, how should the proper value be imposed to the momentum $m_{t}$? }
We suggest that the momentum which have accumulated the gradient from a \textit{single epoch} of FP training is enough to control the gradient approximation error that occurs in every training pipeline. This strategy, called \textit{StatAssist}, gives another answer to control the instability in the initial stage of the training; while previous QAT focuses on a good pre-trained weight, StatAssist focuses on a good initialization of the momentum.

\subsection{Stochastic Gradient Boosting}
\label{sec:gradboost}
Even with the proposed momentum initialization, \textit{StatAssist}, there still exists a possibility of early-convergence due to the gradient instability caused by the inaccurate initial quantization statistics. The gradient calculated with erroneous information may narrow the search space for optimal local-minima and drop the performance. Previous works~\cite{jacob2018quantization,DBLP:journals/corr/abs-1806-08342} suggest to postpone activation quantization for certain extent or use the pre-trained weight of FP model to walkaround this issue. 

We suggest a simple modification to the weight update mechanism in Equation~\ref{eq:gradient_update_approx} to get over the unexpected local-minima in early stages of QAT. In each training step, the gradient $g(W_{t,q})$ is computed using STE during the back-propagation. Our stochastic gradient boosting, \textit{GradBoost} works as follows: 

In each update step $t$, We first define a probability distribution of quantized weights $\Psi(g(W_q))$. Among various probability distributions, we chose a Laplace distribution
$Laplace(0,b_{W,t})$ with a scale parameter $b_{W,t}$ by layer-wise analysis of the histogram of gradients~(Figure~\ref{fig:1}). 

\noindent In each update step $t$, the term $b_{W,t}$ is updated as follows:
\begin{equation}
\label{eq:sensitivity_update}
b_{W,t} = EM^{max}_{t}(g(W_q)) - EM^{min}_{t}(g(W_q))
\end{equation}
where $EM^{max}_{t}(g(W_q))$ is the exponential moving average of $max(g(W_q))$ and $EM^{min}_{t}(g(W_q))$ is the exponential moving average of $min(g(W_q))$ in each update step $t$. 

We further choose a random subset of weights $D_t$ from $W_{t,q}$. For each $w \in D_t$, we apply some distortion to its gradient $g(w)$ with $\psi \sim \Psi(g(W_q))$ in a following way:
\begin{equation}
\label{eq:noise_abs}
\psi \leftarrow sign(g(w)) * |\psi|
\end{equation}
\begin{equation}
\label{eq:noise_clamp}
\psi \leftarrow min(max(\psi,0),\epsilon) 
\end{equation}
\begin{equation}
\label{eq:grad_boost}
g(w) \leftarrow g(w) + \lambda\psi
\end{equation}
where $\epsilon$ is a clamping factor to prevent the exploding gradient problem and $\lambda$ is taken to the power of $t$ for an exponential decay. By matching the sign of $\psi$ with the original gradient $g(w)$ as in Equation~\ref{eq:noise_abs} and adding $\psi$ to $g(w)$ randomly boosts the gradient toward its current direction. For each $w \notin D_t$, the gradient $g(w)$ remains unchanged.  

Note that our GradBoost can be easily combined with Equations~\ref{eq:gradient_update} and ~\ref{eq:gradient_update_approx} and use it as an add-on to any stochastic gradient descent~(SGD) optimizers~\cite{Robbins2007ASA,hinton2012neural, kingma2014adam,DBLP:journals/corr/abs-1711-05101} . 
In our supplementary material, we also provide detailed workflow examples of StatAssist and GradBoost QAT, implemented with Pytorch 1.6~\cite{pytorch}.

\subsection{Optimal \texttt{INT8} QAT from Scratch}
In our supplementary material, we demonstrate the effect of StatAssist and GradBoost on quantized model performances with various lightweight models. Due to its training instability, QAT have required a full-precision~(\texttt{FLOAT32}) pre-trained weight for fine-tuning and the performance is bound to the original \texttt{FLOAT32} model with floating-point computations. From extent experiments, we discovered that the scratch QAT shows comparable performance to the full-precision counterpart without any help of the pre-trained weights, especially when the model becomes complicated.  We also show that our method successfully enables QAT of various deep models from scratch: classification, object detection, semantic segmentation, and style transfer, with comparable or often better performance than their \texttt{FLOAT32} baselines.

\section{FrostNet}
\label{sec:FrostNet}
Using computationally efficient inverted residual bottleneck (\textit{MBConv})~\cite{mobilenetv2} as a building block for hardware-aware NAS, current state-of-the-art mobile-target network architectures \cite{Tan2019MnasNetPN, howard2019searching, Tan2019EfficientNetRM} are showing great performances with practical accuracy-latency trade-offs. While changing the width, depth, and resolution of a network is efficient enough to control the trade-off between accuracy and latency \cite{Tan2019EfficientNetRM}, it's still in the level of macroscopic manipulation. 

While adding cost-efficient add-on modules such as \textit{squeeze-and-excite} (SE) or using different non-linearities like \textit{swish} allows effective microscopic  manipulations in the full-precision (\texttt{FLOAT32}) environment, those are not practical in low-precision (\texttt{INT8}) or mobile settings. Especially for the quantization, it is important to compose a network only with series of \textit{Conv-BN-ReLU}, \textit{Conv-BN}, and \textit{Conv-ReLU} for the best quantized performance by supporting \textit{layer fusion}. 

For the building block of our novel FrostNet, we introduce \textit{Feature Reduction Operation with Squeeze and concaT (Frost)} bottleneck, \textit{FrostConv}, which uses newly designed SE-like module as an add-on for the \textit{MBConv} block. 

\subsection{NAS Based Block Design}
\label{sec:pruning_process}
Before we start block design, we first used the gradient-based NAS, PDARTS \cite{chen2019progressive}, with StatAssist and GradBoost QAT setting to find promising candidates that can replace the SE module. Using squeeze convolution \cite{iandola2016squeezenet} (\textit{SqueezeConv}), expand convolution \cite{iandola2016squeezenet, mobilenetv2} (\textit{ExpandConv}), average pooling (\textit{AvgPool}), and max pooling (\textit{MaxPool}) as cell component candidates, we trained PDARTS with CIFAR10 and CIFAR100 \cite{cifar10} datasets. The cells we initially found are depicted in our supplementary material. 

Since initial cells found with PDARTS were still too computationally expensive, we also performed manual pruning process with following criteria:
\begin{itemize}
    \item Remove redundant concatenations (\textit{Concat}) to reduce memory access during CPU inference.
    \item Only use MaxPool because AvgPool tends to cause errors in QAT setting.
    \item Remove the ${k-2}^{th}$ branch from cells.
    \item Select the most dominant branch as a main convolution module.
    \item Reuse ${k-1}^{th}$ features by concatenating them with ${k}^{th}$ features to reduce size of the model and efficiently reuse feature maps.
    \item Fix the minimum channel size to 8 since channel size under 8 slows down operations.
\end{itemize}

\begin{figure}[!htb]
  \centering
\includegraphics[width=0.6\linewidth]{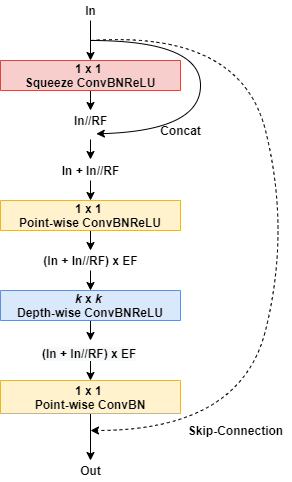}
\vspace{-3mm}
\caption{
Proposed Frost bottleneck (\textit{FrostConv}). 
\textbf{In} : Input channel
size.
\textbf{Out} : Output channel size.
\textbf{EF} : Channel expansion factor. Expands $n$ channels to $n \times EF$. 
\textbf{RF} : Channel squeeze factor. Squeezes $n$ channels to $n // RF$. 
\textbf{ConvBN} : \textit{Conv2d - BatchNorm}.
\textbf{ConvBNReLU} : \textit{Conv2d - BatchNorm - ReLU}.
\textbf{Concat} : \textit{Concatenate} operation.
\textbf{Skip-connection} : Residual skip connection with \textit{Add} operation. Only executed when the input channel size is same as the output channel size and the stride is 2.
%Instead of adding the \textit{squeeze-and-excite} (SE) module to the inverted residual bottleneck (\textit{MBConv}) \cite{Tan2019MnasNetPN,howard2019searching}, 
%We place the \textit{squeeze-and-concatenate} module before the expansion layer.  
}
\label{fig:block_design}
\end{figure}

After small grid search to find some block prototypes, we compared the performance of block candidates with other widely used mobile building blocks using EfficientNet-B0 \cite{Tan2019EfficientNetRM} as a base architecture and chose the one with best overall performance as our Frost bottleneck block (\textit{FrostConv}). 
%The final design of \textit{FrostConv} is shown in Figure \ref{fig:block_design}.
Figure \ref{fig:block_design} shows the final \textit{FrostConv} architecture.

\subsection{Block Specification}
Our Frost bottleneck (\textit{FrostConv}) is based on inverted residual bottleneck (\textit{MBConv}) \cite{mobilenetv2} . To replace the SE \cite{Hu2018SqueezeandExcitationN} module which is usually placed after the depth-wise convolution filters, we place $1 \times 1$ channel squeeze convolution that reduces the channel size by a certain reduce factor (RF) before $1 \times 1$ channel expansion convolution. After concatenating squeezed features with raw output features from the previous block, we passed them through the expansion convolution. This squeeze-and-concatenate (SC) module acts as a quantization-friendly SE module since the SC module fully supports \textit{layer fusion}. Our \textit{FrostConv} shows average $65\%$ latency reduction rate when quantized while the MBConv with SE module shows only $46.63\%$ with similar accuracy performance in classification tasks.

\subsection{Architecture Search}
We employ proxylessNAS \cite{cai2018proxylessnas} with StatAssist and GradBoost QAT setting and reinforcement learning (RL) based objective function using our \textit{FrostConv} as the building block. We also use multi-objective reward $ACC(m) \times {[FLOPS(m)/TAR]}^{w}$ to approximate Pareto-optimal solutions, by balancing model accuracy $ACC(m)$ and FLOPs $FLOPS(m)$ for each model $m$ on the target FLOPs $TAR$. Specifically, we use $3 \times 3$ and $5 \times 5$ depth-wise convolution \cite{howard2017mobilenets}, $\times 3$ and $\times 6$ channel expansion factor (EF), and $\times0.25$ and $\times0.5$  channel squeeze factor (RF) as a set of block configuration. For target FLOPs, we use 500M (Large), 400M (Base), and 300M (Small) FLOPs in search for high to low resource-targeted models respectively.

\begin{table}[t]
\centering

  \begin{tabular}{lccccc}
    \toprule
    In & Operation & Out & EF & RF & s\\
    \midrule
    $224^2 \times 3$  & ConvBNReLU, 3x3 & 32  & - & - & 2\\
    
    $112^2 \times 32$ & FrostConv,    3x3 & 16  & 1 & 1 & 1\\
    $112^2 \times 16$ & FrostConv,    5x5 & 24  & 6 & 4 & 2\\
    $56^2 \times  24$ & FrostConv,    3x3 & 24  & 3 & 4 & 1\\
    
    $56^2 \times  24$ & FrostConv,    5x5 & 40  & 3 & 4 & 2\\ 
    $28^2 \times 40$  & FrostConv,    5x5 & 40  & 3 & 4 & 1\\ 
    
    $28^2 \times 40$  & FrostConv,    5x5 & 80  & 3 & 4 & 2\\ 
    $14^2 \times 80$  & FrostConv,    3x3 & 80  & 3 & 4 & 1\\ 
    
    $14^2 \times 80$  & FrostConv,    5x5 & 96  & 3 & 2 & 1\\ 
    $14^2 \times 96$  & FrostConv,    3x3 & 96  & 3 & 4 & 1\\ 
    $14^2 \times 96$  & FrostConv,    5x5 & 96  & 3 & 4 & 1\\ 
    $14^2 \times 96$  & FrostConv,    5x5 & 96  & 3 & 4 & 1\\ 
    
    $14^2 \times 96$  & FrostConv,    5x5 & 192 & 6 & 2 & 2\\ 
    $7^2 \times 192$  & FrostConv,    5x5 & 192 & 3 & 2 & 1\\ 
    $7^2 \times 192$  & FrostConv,    5x5 & 192 & 3 & 2 & 1\\    
    $7^2 \times 192$  & FrostConv,    5x5 & 192 & 3 & 2 & 1\\  
    
    $7^2 \times 192$  & FrostConv,    5x5 & 320 & 6 & 2 & 1\\ 
    
    $7^2 \times 192$  & ConvBNReLU, 1x1 & 1280& - & - & 1\\  
    $7^2 \times 1280$ & AvgPool,    7x7 &  -  & - & - & 1\\      
    $1^2 \times 1280$ & Conv2d,     1x1 & $k$ & - & - & 1\\       
    \bottomrule
  \end{tabular}
\vspace{-3mm}
\caption{Specifications for FrostNet-Base. $k$ denotes number of class labels (1000).
\textbf{In} : Input resolution and channel size.
\textbf{Out} : Output channel size.
\textbf{EF} : Channel expansion factor. Expands $n$ channels to $n \times EF$. 
\textbf{RF} : Channel squeeze factor. Squeezes $n$ channels to $n // RF$. 
\textbf{s} : Stride.
\textbf{Conv2d} : 2d convolution filter.
\textbf{ConvBNReLU} : \textit{Conv2d - BatchNorm - ReLU}.
\textbf{FrostConv} : Frost bottleneck.
\textbf{AvgPool} : 2d average pooling.
}
\vspace{-2mm}
\label{table:frost_arch}
\end{table}

\begin{table*}[!htb]
\centering

\begin{tabular}{lccccccc}
\toprule
Model                                            & Params & FLOPs & Top-1(f) & Top-1(q) & CPU(f) & CPU(q) & \\ 
%MobileNetV2 1.0 ~\cite{mobilenetv2}                  & 3.4 M  & 300 M & 72.0 & 70.9 & 190 ms & 57 ms &\\
%MobileNetV2 0.35 ~\cite{mobilenetv2}                 & 1.6 M  &  59 M & 60.8 & 57.2 & 109 ms & 38 ms &\\

\midrule
GhostNet 0.5           ~\cite{han2020ghostnet}        & 2.6 M  &  42 M & 66.2 &  -   & 197 ms &433 ms &\\
MobileNetV3-Small 0.75 ~\cite{howard2019searching}    & 2.4 M  &  44 M & 65.4 &  -   & 107 ms & 52 ms &\\
FrostNet-Small 0.5                                    & 2.3 M  & 100 M & \textbf{65.8} & \textbf{63.1} & 125 ms & \textbf{43 ms} &\\
FrostNet-Base 0.5                                     & 2.3 M  & 112 M & \textbf{68.7} & \textbf{66.0} & 137 ms & \textbf{49 ms} &\\
FrostNet-Large 0.5                                    & 2.6 M  & 141 M & \textbf{69.8} & \textbf{67.7} & 151 ms & \textbf{53 ms} &\\
\midrule
MobileNetV3-Small 1.0  ~\cite{howard2019searching}    & 2.7 M  &  54 M & 67.4 & 64.9 & 113 ms & 57 ms &\\
MobileNetV3-Small 1.25 ~\cite{howard2019searching}    & 3.2 M  &  96 M & 70.4 &  -   & 129 ms & 61 ms &\\
FrostNet-Small 0.75                                   & 3.4 M  & 211 M & \textbf{72.6} & \textbf{71.2} & 158 ms & \textbf{58 ms} &\\

\midrule

GhostNet 1.0           ~\cite{han2020ghostnet}        & 5.2 M  & 141 M & 73.9 &  -   & 229 ms &381 ms &\\
MobileNetV3-Large 0.75 ~\cite{howard2019searching}    & 4.0 M  & 155 M & 73.3 &  -   & 176 ms & 74 ms &\\
FrostNet-Small 1.0                                    & 4.8 M  & 315 M & \textbf{74.6} & \textbf{72.3} & 185 ms & \textbf{62 ms} &\\
FrostNet-Large 0.75                                   & 4.0 M  & 302 M & \textbf{74.9} & \textbf{73.5} & 200 ms & \textbf{67 ms} & \\
\midrule

GhostNet 1.3          ~\cite{han2020ghostnet}         & 7.3 M  & 226 M & 75.7 &  -   & 298 ms &773 ms &\\
MobileNetV3-Large 1.0  ~\cite{howard2019searching}    & 5.4 M  & 219 M & 75.2 & 73.8 & 197 ms & 80 ms &\\
EfficientNet-Lite0      ~\cite{Tan2019EfficientNetRM} & 4.7 M  & 407 M & 75.1 & 74.4 & 197 ms & 82 ms &\\
FrostNet-Base 1.0                                     & 5.0 M  & 362 M & \textbf{75.6} & \textbf{74.9} & 233 ms & \textbf{70 ms} &\\
FrostNet-Large 1.0                                    & 5.8 M  & 449 M & \textbf{76.5} & \textbf{75.6} & 234 ms & \textbf{78 ms} &\\
\midrule

MobileNetV3-Large 1.25  ~\cite{howard2019searching}   & 7.5 M  & 373 M & 76.6 &  -   & 240 ms & 98 ms &\\
EfficientNet-Lite1      ~\cite{Tan2019EfficientNetRM} & 5.4 M  & 631 M & 76.7 & 75.9 & 273 ms &109 ms &\\
FrostNet-Base 1.25                                    & 6.9 M  & 582 M & \textbf{77.1} & \textbf{76.5} & 273 ms & \textbf{97 ms} & \\
\midrule

EfficientNet-Lite2      ~\cite{Tan2019EfficientNetRM} & 6.1 M  & 899 M & 77.6 & 77.0 & 350 ms &133 ms &\\
FrostNet-Large 1.25                                   & 8.2 M  & 717 M &\textbf{ 77.8} & \textbf{77.0} & 297 ms &\textbf{100 ms} &\\
\midrule
\end{tabular}
\caption
{
Classification results~(Top 1 accuracy) on the ImageNet~\cite{russakovsky2015imagenet} with \texttt{INT8} post-quantization. Models with similar Top-1 accuracy and CPU latency are grouped together for efficient comparison. \textbf{Params}: Number of parameters of each model. 
\textbf{FLOPs}: FLOPs of each model with different width multipliers. 
\textbf{Top-1(f)}: Top-1 classification accuracy in \texttt{FLOAT32} 
\textbf{Top-1(q)}: Top-1 classification accuracy in \texttt{INT8}. 
\textbf{CPU(f)}: CPU latency in \texttt{FLOAT32} 
\textbf{CPU(q)}: CPU latency in \texttt{INT8}.  
In each group, our FrostNets consistently show higher Top-1 accuracy in both full-precision and post-quantization settings with less quantized CPU latency.
}
\label{table:frost_classification}
\end{table*}
\subsection{Architecture Specification}
The specification for \textit{FrostNet-Base} architecture is provided in Table \ref{table:frost_arch}. Specifications of other FrostNet architectures (\textit{Large, Small}) are also included in our supplementary material. FrostNet-Large and FrostNet-Small are targeted at high and low computing resource environments respectively. For optimized CPU and GPU performance, the minimum output channel size for each convolution is set to 8. 

Furthermore, we have used \textit{channel width multipliers} 0.5, 0.75, 1.0, and 1.25 with a fixed resolution of 224 to cover broader range of classification performance (Top 1 accuracy) in our experiments. We fixed the stem and head while scaling models up to keep them small and fast.

\begin{table}[!htb]
\centering

\begin{subtable}{\linewidth}
  \caption{RetinaNet \cite{lin2018focal}}
  \centering
\begin{tabular}{lccc}
\toprule
Backbone & FLOPs & mAP   &  \\ 
\midrule
MobileNetV2 1.0 ~\cite{mobilenetv2,han2020ghostnet}             & 300 M & 26.7 \\
MobileNetV3-Large 1.0 ~\cite{howard2019searching, han2020ghostnet}   & 219 M & 26.4 \\
GhostNet    1.1 ~\cite{han2020ghostnet}                         & 164 M & 26.6 \\
EfficientNet-Lite0      ~\cite{Tan2019EfficientNetRM}           & 407 M & 29.2\\
FrostNet-Small    1.0                                            & 315 M & \textbf{33.7} \\
FrostNet-Base    1.0                                             & 362 M & \textbf{35.6} \\
FrostNet-Large    1.0                                             & 449 M & \textbf{36.5} \\
\midrule
ResNet18 ~\cite{he2016deep}                                      & 1.8 B & 31.8 \\
ResNet34 ~\cite{he2016deep}                                      & 3.6 B & 34.3 \\
ResNet50 ~\cite{he2016deep}                                      & 3.8 B & 35.8 \\
ResNet101 ~\cite{he2016deep}                                     & 7.6 B & 37.6 \\
\midrule
\end{tabular}
\end{subtable}

\begin{subtable}{\linewidth}
  \caption{Faster R-CNN \cite{ren2015faster}}
  \centering
\begin{tabular}{lccc}
\toprule
Backbone & FLOPs & mAP   &  \\ 
\midrule
MobileNetV2 1.0 ~\cite{mobilenetv2,han2020ghostnet}             & 300 M & 27.5 \\
MobileNetV3-Large 1.0 ~\cite{howard2019searching, han2020ghostnet}  & 219 M & 26.9 \\
GhostNet    1.1 ~\cite{han2020ghostnet}                         & 164 M & 26.9 \\
EfficientNet-Lite0      ~\cite{Tan2019EfficientNetRM}           & 407 M & 31.8\\
FrostNet-Small    1.0                                            & 315 M & \textbf{34.4} \\
FrostNet-Base    1.0                                            & 362 M & \textbf{36.1} \\
FrostNet-Large    1.0                                            & 449 M & \textbf{37.1} \\
\midrule
ResNet18 ~\cite{he2016deep}                                     & 1.8 B & 33.4 \\
ResNet34 ~\cite{he2016deep}                                     & 3.6 B & 36.8 \\
ResNet50 ~\cite{he2016deep}                                      & 3.8 B & 38.4 \\
ResNet101 ~\cite{he2016deep}                                     & 7.6 B & 39.8 \\
\midrule
\end{tabular}
\end{subtable}
\caption
{
Object detection results~(mAP) on MS COCO 2017 \cite{lin2015microsoft} \textit{val} split with one-stage RetinaNet \cite{lin2018focal} and two-stage Faster R-CNN \cite{ren2015faster} with Feature Pyramid Networks (FPN) \cite{ren2015faster, lin2017feature}. \textbf{Backbone}: Type of ImageNet \cite{imagenet_cvpr09} pre-trained backbone. 
\textbf{Detection Framework}: Type of detection framework with FPN. 
\textbf{Params}: Number of parameters of each backbone model. 
\textbf{FLOPs}: FLOPs of each backbone model measured w.r.t. a $224 \times 224$ input. Our FrostNets show significant performances (mAP) as the backbone feature extractor compared to other lightweight models.
}
\label{table:frost_obj_detection}
\end{table}

\section{Experiments}
\label{sec:experiments}

In this section, we demonstrate the practical and cost-efficient performance of our proposed FrostNet with results on classification and object detection tasks. Performance comparison with other state-of-the-art quantization-friendly lightweight models are provided in Table~\ref{table:frost_classification}. We also show effectiveness of our FrostNet as a drop-in replacement for the backbone feature extractor of object detection frameworks in Table~\ref{table:frost_obj_detection}.

\subsection{Experimental Setting}
For classification, we use ImageNet \cite{imagenet_cvpr09} as a benchmark dataset and compare accuracy versus latency with other state-of-the-art lightweight models in both \texttt{FLOAT32} and post-quantized setting. We further evaluate the performance of our model as a feature extractor for well-known object detection models, RetinaNet \cite{lin2018focal} and Faster R-CNN \cite{ren2015faster} with Feature Pyramid Networks (FPN) \cite{ren2015faster, lin2017feature}, on MS COCO object detection benchmark \cite{lin2015microsoft}. We train our classification models with newly proposed training setting in \cite{han2020rexnet}. Using pre-trained FrostNet weights as feature extractor backbones, we train RetinaNet and Faster R-CNN with the original training settings from \cite{lin2018focal} and \cite{ren2015faster}. 

For fair evaluation without any hardware-specific acceleration, we measure the latency of each model using a single machine equipped with AMD Ryzen Threadripper 3960X 24-Core Processor, 2 NVIDIA RTX Titan GPU cards, and Pytorch 1.6 \cite{pytorch}. We use 4 threads to measure \texttt{FLOAT32} and quantized \texttt{INT8} latency on CPU.

\subsection{Classification}
Table \ref{table:frost_classification} shows the performance of all FrostNet models on ImageNet \cite{imagenet_cvpr09} classification task with other state-of-the-art lightweight models. We have also included the performance of each model in \texttt{FLOAT32} full-precision setting for a rough analysis of models with no officially reported post-quantization accuracy results. Our models outperform the current state of the art mobile and edge device target lightweight models such as GhostNet \cite{han2020ghostnet}, MobileNetV3 \cite{howard2019searching}, and EfficientNet-Lite \cite{Tan2019EfficientNetRM}. In each group, our FrostNets consistently show higher Top-1 accuracy in both full-precision and post-quantization settings with less quantized CPU latency.   

Components of a network have a great influence on the latency compression rate when quantized. As in Table \ref{table:quant_latency}, widely-used squeeze-and-excite (SE) \cite{Hu2018SqueezeandExcitationN} module and \textit{swish} \cite{ramachandran2017searching, howard2019searching,Tan2019EfficientNetRM} are not suitable for post-quantization. Modules like channel-shuffle \cite{zhang2017shufflenet} and Ghost \cite{han2020ghostnet} even show counter effects in quantized setting with increased latency. In contrast, our FrostNet models show better accuracy-latency trade-offs in the \texttt{INT8} quantized setting. With \texttt{INT8} post-quantization, our models show average 65\% reduced CPU latency. 

Exponential activation functions (i.e., \textit{sigmoid}, \textit{swish}) force the lower-bit (\texttt{INT8}) to full-precision (\texttt{FLOAT32}) conversion for the exponential calculation, leading to a significant latency drop. The use of a hard-approximation version (i.e., \textit{hard-sigmoid}, \textit{hard-swish}) \cite{howard2019searching} can be a walkaround, but not optimal solution for quantization-friendly model modification. For the optimal quantization performance,  we provide specific tips on network architecture modification for better trade-off between accuracy (mAP, mIOU) and computational resource usage~(latency, weight file size) in our supplementary material.

\subsection{Object Detection}
For further evaluation on the generalized feature extraction ability of our model, we replace the backbone of RetinaNet \cite{lin2018focal} and Faster R-CNN \cite{ren2015faster} with ImageNet pre-trained FrostNets and conduct object detection experiments on MS COCO 2017 \cite{lin2015microsoft}. While maintaining computational resource usages measured in Table \ref{table:frost_obj_detection}, our FrostNets achieve outperforming mean Average Precision (mAP) on MS COCO 2017 \textit{val} split with both one-stage RetinaNet and two-stage Faster R-CNN Feature Pyramid Networks (FPN) \cite{ren2015faster, lin2017feature} as in Table \ref{table:frost_obj_detection}.  

While other lightweight models show insufficient trade-offs between the mAP and backbone FLOPs, our models achieve similar mAP with ResNet~\cite{he2016deep} backbones with significantly reduced FLOPs. Results in Table~\ref{table:frost_obj_detection} demonstrate that FrostNet can act as a drop-in replacement for the backbone feature extractor to deploy various object detection frameworks on mobile devices or other resource limited environments. 
\section{Conclusion}
\label{sec:conclusion}
In this paper, we present a series of approaches towards quantization-aware network architecture search (NAS) to find a network that guarantees both full-precision (\texttt{FLOAT32}) and quantized (\texttt{INT8}) performances. To train NAS algorithms from scratch on the \textit{fake-quantized} environment, we first propose a critical but straightforward optimization method, \textit{StatAssist} and \textit{GradBoost} for stable quantization-aware training from scratch. 
By integrating StatAssist and GradBoost with PDARTS~\cite{chen2019progressive} and proxylessNAS~\cite{cai2018proxylessnas}, we discovered a quantization-friendly network architecture, \textit{FrostNet}.
While conducting better architecture search with our \textit{Frost bottleneck} (\textit{FrostConv}) still remains as an open question, our FrostNet takes a first positive step in quantization-aware architecture search.       

\section*{Acknowledgement}
We would like to thank Clova AI Research team, especially Jung-Woo Ha for their helpful feedback and discussion. NAVER Smart Machine Learning (NSML) platform~\cite{NSML} has been used in the experiments.

{\small
\bibliographystyle{ieee_fullname}
\bibliography{main}

\begin{thebibliography}{10}\itemsep=-1pt

\bibitem{DBLP:journals/corr/BengioLC13}
Yoshua Bengio, Nicholas L{\'{e}}onard, and Aaron~C. Courville.
\newblock Estimating or propagating gradients through stochastic neurons for
  conditional computation.
\newblock {\em CoRR}, abs/1308.3432, 2013.

\bibitem{cai2018proxylessnas}
Han Cai, Ligeng Zhu, and Song Han.
\newblock Proxyless{NAS}: Direct neural architecture search on target task and
  hardware.
\newblock In {\em International Conference on Learning Representations}, 2019.

\bibitem{DBLP:journals/corr/abs-1802-02611}
Liang{-}Chieh Chen, Yukun Zhu, George Papandreou, Florian Schroff, and Hartwig
  Adam.
\newblock Encoder-decoder with atrous separable convolution for semantic image
  segmentation.
\newblock {\em CoRR}, abs/1802.02611, 2018.

\bibitem{chen2019progressive}
Xin Chen, Lingxi Xie, Jun Wu, and Qi Tian.
\newblock Progressive differentiable architecture search: Bridging the depth
  gap between search and evaluation.
\newblock In {\em Proceedings of the IEEE International Conference on Computer
  Vision}, pages 1294--1303, 2019.

\bibitem{Cordts2016Cityscapes}
Marius Cordts, Mohamed Omran, Sebastian Ramos, Timo Rehfeld, Markus Enzweiler,
  Rodrigo Benenson, Uwe Franke, Stefan Roth, and Bernt Schiele.
\newblock The cityscapes dataset for semantic urban scene understanding.
\newblock In {\em Proc. of the IEEE Conference on Computer Vision and Pattern
  Recognition (CVPR)}, 2016.

\bibitem{imagenet_cvpr09}
J. Deng, W. Dong, R. Socher, L.-J. Li, K. Li, and L. Fei-Fei.
\newblock {ImageNet: A Large-Scale Hierarchical Image Database}.
\newblock In {\em CVPR}, 2009.

\bibitem{pascal-voc-2007}
M. Everingham, L. Van~Gool, C.~K.~I. Williams, J. Winn, and A. Zisserman.
\newblock The {PASCAL} {V}isual {O}bject {C}lasses {C}hallenge 2007 {(VOC2007)}
  {R}esults.
\newblock
  http://www.pascal-network.org/challenges/VOC/voc2007/workshop/index.html.

\bibitem{fan2020training}
Angela Fan, Pierre Stock, Benjamin Graham, Edouard Grave, Remi Gribonval, Herve
  Jegou, and Armand Joulin.
\newblock Training with quantization noise for extreme fixed-point compression.
\newblock {\em arXiv preprint arXiv:2004.07320}, 2020.

\bibitem{pmlr-v15-glorot11a}
Xavier Glorot, Antoine Bordes, and Yoshua Bengio.
\newblock Deep sparse rectifier neural networks.
\newblock In Geoffrey Gordon, David Dunson, and Miroslav Dudík, editors, {\em
  Proceedings of the Fourteenth International Conference on Artificial
  Intelligence and Statistics}, volume~15 of {\em Proceedings of Machine
  Learning Research}, pages 315--323, Fort Lauderdale, FL, USA, 2011. PMLR.

\bibitem{NIPS2014_5423}
Ian Goodfellow, Jean Pouget-Abadie, Mehdi Mirza, Bing Xu, David Warde-Farley,
  Sherjil Ozair, Aaron Courville, and Yoshua Bengio.
\newblock Generative adversarial nets.
\newblock In Z. Ghahramani, M. Welling, C. Cortes, N.~D. Lawrence, and K.~Q.
  Weinberger, editors, {\em Advances in Neural Information Processing Systems
  27}, pages 2672--2680. Curran Associates, Inc., 2014.

\bibitem{han2020rexnet}
Dongyoon Han, Sangdoo Yun, Byeongho Heo, and YoungJoon Yoo.
\newblock Rexnet: Diminishing representational bottleneck on convolutional
  neural network, 2020.

\bibitem{han2020ghostnet}
Kai Han, Yunhe Wang, Qi Tian, Jianyuan Guo, Chunjing Xu, and Chang Xu.
\newblock Ghostnet: More features from cheap operations, 2020.

\bibitem{he2016deep}
Kaiming He, Xiangyu Zhang, Shaoqing Ren, and Jian Sun.
\newblock Deep residual learning for image recognition.
\newblock In {\em Proceedings of the IEEE conference on computer vision and
  pattern recognition}, pages 770--778, 2016.

\bibitem{hinton2012neural}
Geoffrey Hinton, Nitish Srivastava, and Kevin Swersky.
\newblock lecture 6a overview of mini-batch gradient descent.
\newblock {\em Neural networks for machine learning}, 14(8), 2012.

\bibitem{hinton2015distilling}
Geoffrey Hinton, Oriol Vinyals, and Jeff Dean.
\newblock Distilling the knowledge in a neural network.
\newblock {\em arXiv preprint arXiv:1503.02531}, 2015.

\bibitem{grad_vanishing}
Sepp Hochreiter.
\newblock The vanishing gradient problem during learning recurrent neural nets
  and problem solutions.
\newblock {\em Int. J. Uncertain. Fuzziness Knowl.-Based Syst.},
  6(2):107–116, Apr. 1998.

\bibitem{howard2019searching}
Andrew Howard, Mark Sandler, Grace Chu, Liang-Chieh Chen, Bo Chen, Mingxing
  Tan, Weijun Wang, Yukun Zhu, Ruoming Pang, Vijay Vasudevan, et~al.
\newblock Searching for mobilenetv3.
\newblock In {\em Proceedings of the IEEE International Conference on Computer
  Vision}, pages 1314--1324, 2019.

\bibitem{howard2017mobilenets}
Andrew~G Howard, Menglong Zhu, Bo Chen, Dmitry Kalenichenko, Weijun Wang,
  Tobias Weyand, Marco Andreetto, and Hartwig Adam.
\newblock Mobilenets: Efficient convolutional neural networks for mobile vision
  applications.
\newblock {\em arXiv preprint arXiv:1704.04861}, 2017.

\bibitem{Hu2018SqueezeandExcitationN}
Jie Hu, L. Shen, and Gang Sun.
\newblock Squeeze-and-excitation networks.
\newblock {\em 2018 IEEE/CVF Conference on Computer Vision and Pattern
  Recognition}, pages 7132--7141, 2018.

\bibitem{iandola2016squeezenet}
Forrest~N Iandola, Song Han, Matthew~W Moskewicz, Khalid Ashraf, William~J
  Dally, and Kurt Keutzer.
\newblock Squeezenet: Alexnet-level accuracy with 50x fewer parameters and< 0.5
  mb model size.
\newblock {\em arXiv preprint arXiv:1602.07360}, 2016.

\bibitem{ioffe2015batch}
Sergey Ioffe and Christian Szegedy.
\newblock Batch normalization: Accelerating deep network training by reducing
  internal covariate shift.
\newblock {\em arXiv preprint arXiv:1502.03167}, 2015.

\bibitem{pix2pix}
Phillip Isola, Jun{-}Yan Zhu, Tinghui Zhou, and Alexei~A. Efros.
\newblock Image-to-image translation with conditional adversarial networks.
\newblock {\em CoRR}, abs/1611.07004, 2016.

\bibitem{jacob2018quantization}
Benoit Jacob, Skirmantas Kligys, Bo Chen, Menglong Zhu, Matthew Tang, Andrew
  Howard, Hartwig Adam, and Dmitry Kalenichenko.
\newblock Quantization and training of neural networks for efficient
  integer-arithmetic-only inference.
\newblock In {\em 2018 IEEE/CVF Conference on Computer Vision and Pattern
  Recognition}, pages 2704--2713. IEEE, 2018.

\bibitem{NSML}
Hanjoo Kim, Minkyu Kim, Dongjoo Seo, Jinwoong Kim, Heungseok Park, Soeun Park,
  Hyunwoo Jo, KyungHyun Kim, Youngil Yang, Youngkwan Kim, et~al.
\newblock Nsml: Meet the mlaas platform with a real-world case study.
\newblock {\em arXiv preprint arXiv:1810.09957}, 2018.

\bibitem{kingma2014adam}
Diederik~P. Kingma and Jimmy Ba.
\newblock Adam: A method for stochastic optimization, 2014.

\bibitem{DBLP:journals/corr/abs-1806-08342}
Raghuraman Krishnamoorthi.
\newblock Quantizing deep convolutional networks for efficient inference: {A}
  whitepaper.
\newblock {\em CoRR}, abs/1806.08342, 2018.

\bibitem{cifar}
Alex Krizhevsky.
\newblock Learning multiple layers of features from tiny images.
\newblock Technical report, Canadian Institute For Advanced Research, 2009.

\bibitem{cifar10}
Alex Krizhevsky, Vinod Nair, and Geoffrey Hinton.
\newblock Cifar-10 (canadian institute for advanced research).

\bibitem{lecun1990optimal}
Yann LeCun, John~S Denker, and Sara~A Solla.
\newblock Optimal brain damage.
\newblock In {\em Advances in neural information processing systems}, pages
  598--605, 1990.

\bibitem{loss_landscape}
Hao Li, Zheng Xu, Gavin Taylor, Christoph Studer, and Tom Goldstein.
\newblock Visualizing the loss landscape of neural nets.
\newblock In S. Bengio, H. Wallach, H. Larochelle, K. Grauman, N. Cesa-Bianchi,
  and R. Garnett, editors, {\em Advances in Neural Information Processing
  Systems 31}, pages 6389--6399. Curran Associates, Inc., 2018.

\bibitem{li2020gan}
Muyang Li, Ji Lin, Yaoyao Ding, Zhijian Liu, Jun-Yan Zhu, and Song Han.
\newblock Gan compression: Efficient architectures for interactive conditional
  gans.
\newblock In {\em Proceedings of the IEEE/CVF Conference on Computer Vision and
  Pattern Recognition}, 2020.

\bibitem{Li2019quantizedObjectDet}
R. {Li}, Y. {Wang}, F. {Liang}, H. {Qin}, J. {Yan}, and R. {Fan}.
\newblock Fully quantized network for object detection.
\newblock In {\em 2019 IEEE/CVF Conference on Computer Vision and Pattern
  Recognition (CVPR)}, pages 2805--2814, 2019.

\bibitem{tdsod}
Yuxi Li, Jiuwei Li, Weiyao Lin, and Jianguo Li.
\newblock Tiny-dsod: Lightweight object detection for resource-restricted
  usages.
\newblock {\em CoRR}, abs/1807.11013, 2018.

\bibitem{lin2017feature}
Tsung-Yi Lin, Piotr Dollár, Ross Girshick, Kaiming He, Bharath Hariharan, and
  Serge Belongie.
\newblock Feature pyramid networks for object detection, 2017.

\bibitem{lin2018focal}
Tsung-Yi Lin, Priya Goyal, Ross Girshick, Kaiming He, and Piotr Dollár.
\newblock Focal loss for dense object detection, 2018.

\bibitem{lin2015microsoft}
Tsung-Yi Lin, Michael Maire, Serge Belongie, Lubomir Bourdev, Ross Girshick,
  James Hays, Pietro Perona, Deva Ramanan, C.~Lawrence Zitnick, and Piotr
  Dollár.
\newblock Microsoft coco: Common objects in context, 2015.

\bibitem{liu2018darts}
Hanxiao Liu, Karen Simonyan, and Yiming Yang.
\newblock Darts: Differentiable architecture search.
\newblock {\em arXiv preprint arXiv:1806.09055}, 2018.

\bibitem{liu2016ssd}
Wei Liu, Dragomir Anguelov, Dumitru Erhan, Christian Szegedy, Scott Reed,
  Cheng-Yang Fu, and Alexander~C Berg.
\newblock Ssd: Single shot multibox detector.
\newblock In {\em European conference on computer vision}, pages 21--37.
  Springer, 2016.

\bibitem{DBLP:journals/corr/abs-1711-05101}
Ilya Loshchilov and Frank Hutter.
\newblock Fixing weight decay regularization in adam.
\newblock {\em CoRR}, abs/1711.05101, 2017.

\bibitem{Ma_2018_ECCV}
Ningning Ma, Xiangyu Zhang, Hai-Tao Zheng, and Jian Sun.
\newblock Shufflenet v2: Practical guidelines for efficient cnn architecture
  design.
\newblock In {\em The European Conference on Computer Vision (ECCV)}, 2018.

\bibitem{mehta2018espnet}
Sachin Mehta, Mohammad Rastegari, Anat Caspi, Linda Shapiro, and Hannaneh
  Hajishirzi.
\newblock Espnet: Efficient spatial pyramid of dilated convolutions for
  semantic segmentation.
\newblock In {\em Proceedings of the European Conference on Computer Vision
  (ECCV)}, pages 552--568, 2018.

\bibitem{mehta2018espnetv2}
Sachin Mehta, Mohammad Rastegari, Linda Shapiro, and Hannaneh Hajishirzi.
\newblock Espnetv2: A light-weight, power efficient, and general purpose
  convolutional neural network.
\newblock In {\em Proceedings of the IEEE conference on computer vision and
  pattern recognition}, 2019.

\bibitem{pytorch}
Adam Paszke, Sam Gross, Francisco Massa, Adam Lerer, James Bradbury, Gregory
  Chanan, Trevor Killeen, Zeming Lin, Natalia Gimelshein, Luca Antiga, Alban
  Desmaison, Andreas Kopf, Edward Yang, Zachary DeVito, Martin Raison, Alykhan
  Tejani, Sasank Chilamkurthy, Benoit Steiner, Lu Fang, Junjie Bai, and Soumith
  Chintala.
\newblock Pytorch: An imperative style, high-performance deep learning library.
\newblock In {\em Advances in Neural Information Processing Systems 32}, pages
  8024--8035. Curran Associates, Inc., 2019.

\bibitem{pham2018efficient}
Hieu Pham, Melody~Y. Guan, Barret Zoph, Quoc~V. Le, and Jeff Dean.
\newblock Efficient neural architecture search via parameter sharing, 2018.

\bibitem{ramachandran2017searching}
Prajit Ramachandran, Barret Zoph, and Quoc~V. Le.
\newblock Searching for activation functions, 2017.

\bibitem{ren2015faster}
Shaoqing Ren, Kaiming He, Ross Girshick, and Jian Sun.
\newblock Faster r-cnn: Towards real-time object detection with region proposal
  networks.
\newblock In {\em Advances in neural information processing systems}, pages
  91--99, 2015.

\bibitem{Robbins2007ASA}
Herbert~E. Robbins.
\newblock A stochastic approximation method.
\newblock {\em Annals of Mathematical Statistics}, 22:400--407, 2007.

\bibitem{russakovsky2015imagenet}
Olga Russakovsky, Jia Deng, Hao Su, Jonathan Krause, Sanjeev Satheesh, Sean Ma,
  Zhiheng Huang, Andrej Karpathy, Aditya Khosla, Michael Bernstein, et~al.
\newblock Imagenet large scale visual recognition challenge.
\newblock {\em International Journal of Computer Vision}, 115(3):211--252,
  2015.

\bibitem{mobilenetv2}
M. {Sandler}, A. {Howard}, M. {Zhu}, A. {Zhmoginov}, and L. {Chen}.
\newblock Mobilenetv2: Inverted residuals and linear bottlenecks.
\newblock In {\em 2018 IEEE/CVF Conference on Computer Vision and Pattern
  Recognition}, pages 4510--4520, 2018.

\bibitem{pmlr-v28-sutskever13}
Ilya Sutskever, James Martens, George Dahl, and Geoffrey Hinton.
\newblock On the importance of initialization and momentum in deep learning.
\newblock In Sanjoy Dasgupta and David McAllester, editors, {\em Proceedings of
  the 30th International Conference on Machine Learning}, volume~28 of {\em
  Proceedings of Machine Learning Research}, pages 1139--1147, Atlanta,
  Georgia, USA, 2013. PMLR.

\bibitem{Tan2019MnasNetPN}
M. Tan, Bo Chen, R. Pang, V. Vasudevan, and Quoc~V. Le.
\newblock Mnasnet: Platform-aware neural architecture search for mobile.
\newblock {\em 2019 IEEE/CVF Conference on Computer Vision and Pattern
  Recognition (CVPR)}, pages 2815--2823, 2019.

\bibitem{Tan2019EfficientNetRM}
M. Tan and Quoc~V. Le.
\newblock Efficientnet: Rethinking model scaling for convolutional neural
  networks.
\newblock {\em ArXiv}, abs/1905.11946, 2019.

\bibitem{xu2020pcdarts}
Yuhui Xu, Lingxi Xie, Xiaopeng Zhang, Xin Chen, Guo-Jun Qi, Qi Tian, and
  Hongkai Xiong.
\newblock {\{}PC{\}}-{\{}darts{\}}: Partial channel connections for
  memory-efficient architecture search.
\newblock In {\em International Conference on Learning Representations}, 2020.

\bibitem{zhang2017shufflenet}
Xiangyu Zhang, Xinyu Zhou, Mengxiao Lin, and Jian Sun.
\newblock Shufflenet: An extremely efficient convolutional neural network for
  mobile devices.
\newblock {\em arXiv preprint arXiv:1707.01083}, 2017.

\bibitem{zoph2018learning}
Barret Zoph, Vijay Vasudevan, Jonathon Shlens, and Quoc~V. Le.
\newblock Learning transferable architectures for scalable image recognition,
  2018.

\end{thebibliography}
}

\clearpage
\appendix

\section{Prototype Cells from PDARTS}
\begin{figure*}[!ht]
  \centering
\begin{subfigure}{0.8\textwidth}
  \centering
\includegraphics[width=\linewidth]{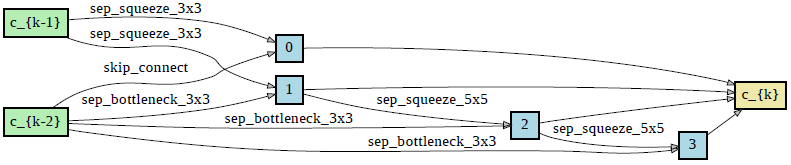}
\caption{CIFAR10 Normal Cell}
\label{fch3_cell_1}
\end{subfigure}
\begin{subfigure}{0.8\textwidth}
  \centering
\includegraphics[width=\linewidth]{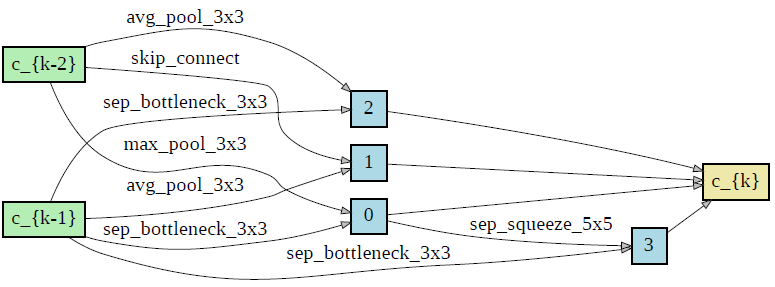}
\caption{CIFAR10 Reduction Cell}
\label{ch3_cell_2}
\end{subfigure}
\begin{subfigure}{0.4\textwidth}
  \centering
\includegraphics[width=\linewidth]{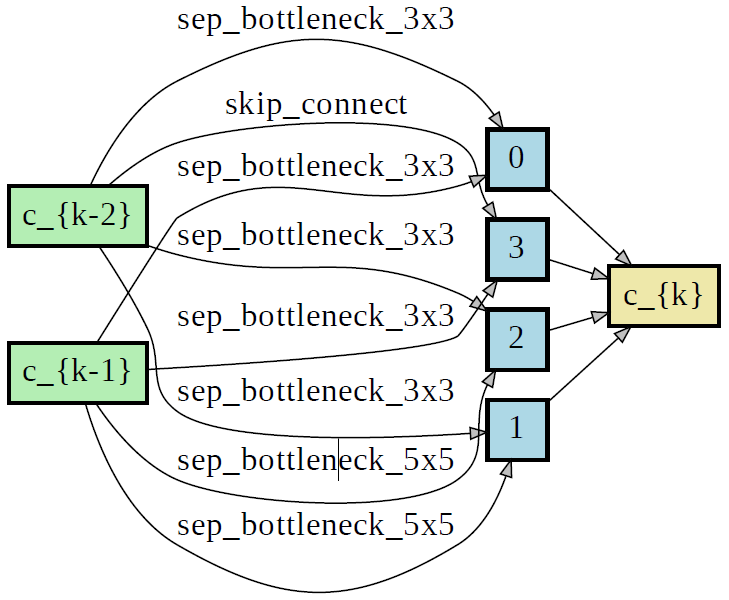}
\caption{CIFAR100 Normal Cell}
\label{ch3_cell3}
\end{subfigure}
\begin{subfigure}{0.4\textwidth}
  \centering
\includegraphics[width=\linewidth]{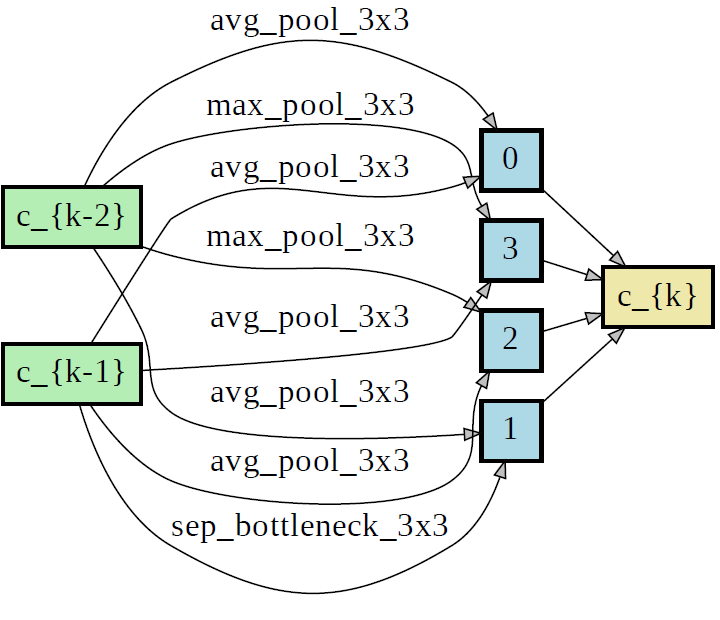}
\caption{CIFAR100 Reduction Cell}
\label{ch3_cell_4}
\end{subfigure}

\caption{
Visualizations of cells found with PDARTS in StatAssist and Gradboost QAT setting. 
\textbf{sep\_squeeze\_nxn}: squeeze convolution with depth-wise convolutions and point-wise convolutions. 
\textbf{sep\_bottleneck\_nxn}: squeeze convolution with $n \times n$ depth-wise convolutions and $1 \times 1$ point-wise convolutions.
\textbf{skip\_connect}: a skip connection with \texttt{Add} operation. 
\textbf{max\_pool\_nxn}: $n \times n$ max pooling. 
\textbf{avg\_pool\_nxn}: $n \times n$ average pooling.
}
\label{ch3_cell}
\end{figure*}

We include cell prototypes found with PDARTS~\cite{chen2019progressive} in Figure~\ref{ch3_cell}. Normal cells are used when stride is 1 and Reduction cells are used when stride is 2.

\section{FrostNet Specifications}

Specifications for FrostNet-Large and FrostNet-Small are provided in Table ~\ref{table:frost_arch_large} and ~\ref{table:frost_arch_small} respectively. FrostNet-Large and FrostNet-Small are targeted at high and low computing resource environments respectively. The models are searched through applying proxylessNAS \cite{cai2018proxylessnas} with reinforcement learning (RL) based objective function. The minimum output channel size for each convolution is set to 8 for optimized CPU and GPU performance. 

\begin{table}[!ht]
\centering
  \begin{tabular}{lccccc}
    \toprule
    In & Operation & Out & EF & RF & s\\
    \midrule
    $224^2 \times 3$  & ConvBNReLU, 3x3 & 32  & - & - & 2\\
    
    $112^2 \times 32$ & FrostConv,    3x3 & 16  & 1 & 1 & 1\\
    $112^2 \times 16$ & FrostConv,    3x3 & 24  & 6 & 4 & 2\\
    $56^2 \times  24$ & FrostConv,    3x3 & 24  & 3 & 4 & 1\\
    
    $56^2 \times  24$ & FrostConv,    5x5 & 40  & 6 & 4 & 2\\ 
    $28^2 \times 40$  & FrostConv,    3x3 & 40  & 3 & 4 & 1\\ 
    
    $28^2 \times 40$  & FrostConv,    5x5 & 80  & 6 & 4 & 2\\ 
    $14^2 \times 80$  & FrostConv,    5x5 & 80  & 3 & 4 & 1\\ 
    $14^2 \times 80$  & FrostConv,    5x5 & 80  & 3 & 4 & 1\\ 
    
    $14^2 \times 80$  & FrostConv,    5x5 & 96  & 6 & 4 & 1\\ 
    $14^2 \times 96$  & FrostConv,    5x5 & 96  & 3 & 4 & 1\\ 
    $14^2 \times 96$  & FrostConv,    3x3 & 96  & 3 & 4 & 1\\ 
    $14^2 \times 96$  & FrostConv,    3x3 & 96  & 3 & 4 & 1\\

    $14^2 \times 96$  & FrostConv,    5x5 & 192 & 6 & 2 & 2\\ 
    $7^2 \times 192$  & FrostConv,    5x5 & 192 & 6 & 4 & 1\\ 
    $7^2 \times 192$  & FrostConv,    5x5 & 192 & 6 & 4 & 1\\    
    $7^2 \times 192$  & FrostConv,    5x5 & 192 & 3 & 4 & 1\\   
    $7^2 \times 192$  & FrostConv,    5x5 & 192 & 3 & 4 & 1\\       
    
    $7^2 \times 192$  & FrostConv,    5x5 & 320 & 6 & 2 & 1\\  
    
    $7^2 \times 192$  & ConvBNReLU, 1x1 & 1280& - & - & 1\\  
    $7^2 \times 1280$ & AvgPool,    7x7 &  -  & - & - & 1\\      
    $1^2 \times 1280$ & Conv2d,     1x1 & $k$ & - & - & 1\\       
    \bottomrule
  \end{tabular}

\caption{Specifications for FrostNet-Large. $k$ denotes number of class labels (1000).
\textbf{In} : Input resolution and channel size.
\textbf{Out} : Output channel size.
\textbf{EF} : Channel expansion factor. Expands $n$ channels to $n \times EF$. 
\textbf{RF} : Channel squeeze factor. Squeezes $n$ channels to $n // RF$. 
\textbf{s} : Stride.
\textbf{Conv2d} : 2d convolution filter.
\textbf{ConvBNReLU} : \textit{Conv2d - BatchNorm - ReLU}.
\textbf{FrostConv} : Frost bottleneck.
\textbf{AvgPool} : 2d average pooling.
}
\label{table:frost_arch_large}
\end{table}

\begin{table}[!ht]
\centering
  \begin{tabular}{lccccc}
    \toprule
    Input & Operation & Out & EF & RF & s\\
    \midrule
    $224^2 \times 3$  & ConvBNReLU, 3x3 & 32  & - & - & 2\\
    $112^2 \times 32$ & FrostConv,    3x3 & 16  & 1 & 1 & 1\\
    $112^2 \times 16$ & FrostConv,    5x5 & 24  & 6 & 4 & 2\\
    $56^2 \times  24$ & FrostConv,    3x3 & 24  & 3 & 4 & 1\\
    
    $56^2 \times  24$ & FrostConv,    5x5 & 40  & 3 & 4 & 2\\ 
    $28^2 \times 40$  & FrostConv,    5x5 & 40  & 3 & 4 & 1\\ 
    
    $28^2 \times 40$  & FrostConv,    5x5 & 80  & 3 & 4 & 2\\ 
    $14^2 \times 80$  & FrostConv,    3x3 & 80  & 3 & 4 & 1\\ 
    
    $14^2 \times 80$  & FrostConv,    5x5 & 96  & 3 & 2 & 1\\ 
    $14^2 \times 96$  & FrostConv,    3x3 & 96  & 3 & 4 & 1\\ 
    $14^2 \times 96$  & FrostConv,    5x5 & 96  & 3 & 4 & 1\\ 
    
    $14^2 \times 96$  & FrostConv,    5x5 & 192 & 6 & 2 & 2\\ 
    $7^2 \times 192$  & FrostConv,    5x5 & 192 & 3 & 2 & 1\\ 
    $7^2 \times 192$  & FrostConv,    5x5 & 192 & 3 & 2 & 1\\   
    
    $7^2 \times 192$  & FrostConv,    5x5 & 320 & 6 & 2 & 1\\  
    
    $7^2 \times 192$  & ConvBNReLU, 1x1 & 1280& - & - & 1\\  
    $7^2 \times 1280$ & AvgPool,    7x7 &  -  & - & - & 1\\      
    $1^2 \times 1280$ & Conv2d,     1x1 & $k$ & - & - & 1\\   
    \bottomrule
  \end{tabular}
\vspace{-2mm}
\caption{Specifications for FrostNet-Small. $k$ denotes number of class labels (1000).
\textbf{In} : Input resolution and channel size.
\textbf{Out} : Output channel size.
\textbf{EF} : Channel expansion factor. Expands $n$ channels to $n \times EF$. 
\textbf{RF} : Channel squeeze factor. Squeezes $n$ channels to $n // RF$. 
\textbf{s} : Stride.
\textbf{Conv2d} : 2d convolution filter.
\textbf{ConvBNReLU} : \textit{Conv2d - BatchNorm - ReLU}.
\textbf{FrostConv} : Frost bottleneck.
\textbf{AvgPool} : 2d average pooling.
}
\label{table:frost_arch_small}
\end{table}

\section{Example Workflow of Quantization-Aware Training}
In this section, we describe an example workflow of our StatAssist and GradBoost quantization-aware training~(QAT) with PyTorch. Our workflow in Algorithm~\ref{alg:qat_workflow} closely follows the methodology of the official PyTorch 1.6 quantization library. Detailed algorithms and PyTorch codes for the StatAssist and Gradboost are also provided in Section~\ref{sp:pytorch_statassist} and \ref{sp:pytorch_gradboost}.

\subsection{StatAssist in Pytorch 1.6}
\label{sp:pytorch_statassist}  
We provide a typical PyTorch 1.6 code illustrating StatAssist implementation in Algorithm~\ref{alg:statassist}. The actual implementation may vary according to training workflows or PyTorch versions.
\begin{algorithm}[!htb]
\caption{Quantization-aware training~(QAT) with StatAssist and GradBoost}
\label{alg:qat_workflow}
\begin{algorithmic}[1]
  \State Prepare a full-precision~(FP) model with \textit{fake-quantization} compatibility.      
  \State Create a training workflow of the Full-precision~(FP) model.
  \State Replace the optimizer with the GradBoost-applied version.  
  \State Run StatAssist using the code provided in Algorithm~\ref{alg:statassist}.
  \State Prepare the model for QAT with ~\textit{layer-fusion} and \textit{fake-quantization}.
  \State Start QAT from scratch.
  \State Convert model to \texttt{INT8}-quantized version.
  \State Evaluate the model performance.
  
  \end{algorithmic}
\end{algorithm}

\begin{algorithm}[!htb]

\caption{PyTorch code for StatAssist}
\label{alg:statassist}
\scriptsize
\begin{lstlisting}[language=Python]
import torch
import torch.nn as nn
import torch.optim as optim
import torch.quantization as quantization

# Define model, optimizer, and learning rate scheduler. 
...

FP_EPOCHS = 1

for epoch in range(FP_EPOCHS):        
    lr = lr_scheduler.step(epoch)

    for param_group in optimizer.param_groups:         
        param_group['lr'] = lr
        train_acc, train_loss = train(model, 
                                    train_loader,  
                                    optimizer, 
                                    criterion,
                                    num_class, epoch, 
                                    device=device)

#prepare the model for quantization-aware training.                                      
model.quantized.fuse_model()
model.quantized.qconfig = \\
        quantization.get_default_qat_qconfig('qnnpack')
quantization.prepare_qat(model, inplace=True) 

# Start training
...

# Convert model to INT8 for evaluation.
quantization.convert(model.eval(),inplace = True) 

\end{lstlisting}
  \end{algorithm}
  
\subsection{GradBoost Optimizers}
\label{sp:pytorch_gradboost}
Our Gradboost method in Section~\ref{sec:gradboost} is applicable to any existing optimizer implementations by adding extra lines to the gradient calculation. An example algorithm for GradBoost-applied momentum-SGD~\cite{pmlr-v28-sutskever13} and AdamW~\cite{DBLP:journals/corr/abs-1711-05101} are provided in Algorithm~\ref{alg:gradboost_sgd} and ~\ref{alg:gradboost_adamw}. Please refer to \texttt{optimizer.py} in our source code for detailed GradBoost applications to Pytorch 1.6 optimizers.

\def\R{{\rm I\hspace{-0.50ex}R}}
\def\E{\mathds{E}}
\def\D{{\mathcal{D}}}
\def\H{\textbf{H}}
\def\A{\cal{A}}
\definecolor{springgreen}{rgb}{0.0, 1.0, 0.5}
\definecolor{thistle}{rgb}{0.85, 0.75, 0.85}
\newcommand{\vc}[1]{\textit{\textbf{#1}}}
\newcommand{\gradboostcolor}{thistle}
\newcommand{\gradboost}[1]{\colorbox{\gradboostcolor}{$\displaystyle #1$}}
\newcommand{\gradboosttext}[1]{\colorbox{\gradboostcolor}{#1}}

\begin{algorithm}[!htb]
\caption{SGD with GradBoost}
\footnotesize
\label{alg:gradboost_sgd}
\begin{algorithmic}[1]
\State{\textbf{given} initial learning rate $\alpha \in \R$, momentum factor $\beta_1 \in \R$, weight decay L$_2$ regularization factor $\lambda \in \R$, exponential-moving max and min decay factor $\gamma_1 \in \R$,noise clamping factor $\gamma_2 \in \R$, and noise decay factor $\gamma_3 \in \R$}
\State{\textbf{initialize} time step $t \leftarrow 0$, parameter vector $\boldmath{\theta}_{t=0} \in \R^n$,  first moment vector $\vc{m}_{t=0} \leftarrow \vc{0}$, schedule multiplier $\eta_{t=0} \in \R$, gradient maximum $max^\vc{g}_{0}\leftarrow 1$, and gradient minimum $min^\vc{g}_{0}\leftarrow 0$}  

\Repeat
	\State{$t \leftarrow t + 1$}
	\State{$\nabla f_t(\boldmath{\theta}_{t-1}) \leftarrow  \text{SelectBatch}(\boldmath{\theta}_{t-1})$}  \Comment{select batch and return the corresponding gradient}
	\State{$\vc{g}_t \leftarrow \nabla f_t(\boldmath{\theta}_{t-1})  + \lambda\boldmath{\theta}_{t-1}$}
	\State{$max^\vc{g}_{t} \leftarrow \gamma_1 max^\vc{g}_{t-1} + (1 - \gamma_1) max(max^\vc{g}_{t-1},\vc{g}_t)$} \Comment{here and below all operations are element-wise}
	\State{$min^\vc{g}_{t} \leftarrow \gamma_1 min^\vc{g}_{t-1} + (1 - \gamma_1) min(min^\vc{g}_{t-1},\vc{g}_t)$}
    \State{$b_{t} = max^\vc{g}_{t} -min^\vc{g}_{t}$}
    \State{Random sample $\psi \sim Laplace(0,b_{t})$ and $k \sim Bernoulli(0,\frac{1}{2})$}
    \State{$\psi \leftarrow sign(\vc{g}_{t}) * |\psi|$}
    \State{$\psi \leftarrow min(max(\psi,0), \gamma_2)$}
    \State{$\vc{g}_t \leftarrow \vc{g}_t + k(1-\gamma^t_3)\psi$} \Comment{add noise only if $k=1$. $\gamma_3$ is taken to the power of $t$}
	\State{$\eta_t \leftarrow \text{SetScheduleMultiplier}(t)$}	\Comment{can be fixed, decay, be used for warm restarts}
	\State{$\vc{m}_t \leftarrow \beta_1 \vc{m}_{t-1} + \eta_t \alpha \vc{g}_t $} \label{sgd-mom1} 
	\State{$\boldmath{\theta}_t \leftarrow \boldmath{\theta}_{t-1} - \vc{m}_t - \eta_t \lambda\boldmath{\theta}_{t-1}$} 
\Until{ \textit{stopping criterion is met} }

\State\Return{optimized parameters $\boldmath{\theta}_t$}
\end{algorithmic}
\end{algorithm}

\begin{algorithm}[!htb]
\caption{AdamW with GradBoost}
\footnotesize
\label{alg:gradboost_adamw}
\begin{algorithmic}[1]
\State{\textbf{given}  initial learning rate $\alpha \in \R$, momentum factor $\beta_1 \in \R$,  second-order momentum factor $\beta_2 \in \R$, $\epsilon = 10^{-8}$,  weight decay L$_2$ regularization factor $\lambda \in \R$, exponential-moving max and min decay factor $\gamma_1 \in \R$,noise clamping factor $\gamma_2 \in \R$, and noise decay factor $\gamma_3 \in \R$}
\State{\textbf{initialize} time step $t \leftarrow 0$, parameter vector $\boldmath{\theta}_{t=0} \in \R^n$,  first moment vector $\vc{m}_{t=0} \leftarrow \vc{0}$, second moment vector  $\vc{v}_{t=0} \leftarrow \vc{0}$, schedule multiplier $\eta_{t=0} \in \R$, gradient maximum $max^\vc{g}_{0}\leftarrow 1$, and gradient minimum $min^\vc{g}_{0}\leftarrow 0$}
\Repeat
	\State{$t \leftarrow t + 1$}
	\State{$\nabla f_t(\boldmath{\theta}_{t-1}) \leftarrow  \text{SelectBatch}(\boldmath{\theta}_{t-1})$}  \Comment{select batch and return the corresponding gradient}
	\State{$\vc{g}_t \leftarrow \nabla f_t(\boldmath{\theta}_{t-1})  + \lambda\boldmath{\theta}_{t-1}$}
	\State{$max^\vc{g}_{t} \leftarrow \gamma_1  max^\vc{g}_{t-1} + (1 - \gamma_1 ) max(max^\vc{g}_{t-1},\vc{g}_t)$} \Comment{here and below all operations are element-wise}
	\State{$min^\vc{g}_{t} \leftarrow \gamma_1  min^\vc{g}_{t-1} + (1 - \gamma_1 ) min(min^\vc{g}_{t-1},\vc{g}_t)$}
    \State{$b_{t} = max^\vc{g}_{t} -min^\vc{g}_{t}$}
    \State{Random sample $\psi \sim Laplace(0,b_{t})$ and $k \sim Bernoulli(0,\frac{1}{2})$}
    \State{$\psi \leftarrow sign(\vc{g}_{t}) * |\psi|$}
    \State{$\psi \leftarrow min(max(\psi,0), \gamma_2) $}
    \State{$\vc{g}_t \leftarrow \vc{g}_t + k(1-\gamma^t_3)\psi$} \Comment{add noise only if $k=1$. $\gamma_3$ is taken to the power of $t$}	
	\State{$\vc{m}_t \leftarrow \beta_1 \vc{m}_{t-1} + (1 - \beta_1) \vc{g}_t $} 
	\State{$\vc{v}_t \leftarrow \beta_2 \vc{v}_{t-1} + (1 - \beta_2) \vc{g}^2_t $}
	\State{$\hat{\vc{m}}_t \leftarrow \vc{m}_t/(1 - \beta_1^t) $} \Comment{$\beta_1$ is taken to the power of $t$} 
	\State{$\hat{\vc{{v}}}_t \leftarrow \vc{v}_t/(1 - \beta_2^t) $} \Comment{$\beta_2$ is taken to the power of $t$} 
	\State{$\eta_t \leftarrow \text{SetScheduleMultiplier}(t)$}	\Comment{can be fixed, decay, or also be used for warm restarts}
	\State{$\boldmath{\theta}_t \leftarrow \boldmath{\theta}_{t-1} - \eta_t \left( \alpha  \hat{\vc{m}}_t / (\sqrt{\hat{\vc{v}}_t} + \epsilon) + \lambda\boldmath{\theta}_{t-1} \right)$} 
\Until{ \textit{stopping criterion is met} }

\State\Return{optimized parameters $\boldmath{\theta}_t$}
\end{algorithmic}
\end{algorithm}

\section{Experiments on StatAssist and GradBoost}

To empirically evaluate our proposed StatAssist and GradBoost, we perform three sets of experiments on training different lightweight models with StatAssist and GradBoost QAT from scratch. The results on classification, object detection, semantic segmentation, and style transfer prove the effectiveness of our method in both quantitative and qualitative ways. 

\subsection{Experimental Setting}
\paragraph{Training Protocol} Our main contribution in Section~\ref{sec:introduction} focuses on making the optimizer robust to gradient approximation error caused by STE during the back-propagation of QAT. To be more specific, we initialize the optimizer with StatAssist and distort a random subset of gradients on each update step via GradBoost. As an optimizer updates its momentum by itself each step, we simply apply StatAssist by running the optimizer with FP model for a single epoch. Our StatAssist also replaces the learning rate warm-up process in conventional model training schemes. For GradBoost, we modify the gradient update step of each optimizer with equations \ref{eq:sensitivity_update} through \ref{eq:grad_boost}. 

\paragraph{Implementation Details}
We train our models using PyTorch~\cite{pytorch} and follow the methodology of PyTorch quantization library. Typical PyTorch~\cite{pytorch} code illustrating StatAssist implementation is in Appendix~\ref{sp:pytorch_statassist}. Detailed algorithms for different GradBoost optimizers are in Appendix~\ref{sp:pytorch_gradboost}. 

For the optimal latency, we tuned the components of each model for the best trade-off between model performance and compression. As mentioned in Section~\ref{sec:static_quant}, the latency gap between the conceptual design and the actual implementation is critical. The \textit{layer fusion}, integrating the convolution, normalization, and activation into a single convolution operation, can improve the latency by reducing the conversion overhead between FP and lower-bit. For better trade-off between the accuracy (mAP, mIOU, image quality) and efficiency~(latency, FLOPs, compression rate), we modified models in the following ways:  
\label{sp:reconstruction}.
\begin{itemize}
    \item We first replaced each normalization and activation function that comes after a convolution~(Conv) layer with the Batch Normalization~(BN)~\cite{ioffe2015batch} and ReLU~\cite{pmlr-v15-glorot11a}. For special modules like \textit{Conv-Concatenate-BN-ReLU} or \textit{Conv-Add-BN-ReLU} in ESPNets~\cite{mehta2018espnet,mehta2018espnetv2}, we inserted an extra \textit{$1\times1$ Conv} before BN. 
    \item For MobileNetV3 + LRASPP, we replaced the \textit{$49\times49$ Avg-Pool Stride=(16, 20)} in LRASPP with \textit{$25\times25$ Avg-Pool Stride=(8, 8)} to train models with $768\times768$ random-cropped images instead of $2048\times1024$ full-scale images.
    \item Quantizing the entire layers of a model except the last single layer yields the best trade-off between accuracy and efficiency. 
\end{itemize}

\begin{table*}[!htb]
\centering
\tabcolsep=0.10cm
\begin{tabular}{@{}lccccccc@{}}
\toprule
\multirow{2}{*}{Model} & 
\multirow{2}{*}{Params} & 
\multirow{2}{*}{FLOPs} & 
\multirow{2}{*}{FP training} &

\multirow{2}{*}{\begin{tabular}[c]{@{}c@{}} QAT\\ Fine-tune \end{tabular}} &
\multirow{2}{*}{\begin{tabular}[c]{@{}c@{}} StatAssist\\ Only\end{tabular}} &
\multirow{2}{*}{\begin{tabular}[c]{@{}c@{}} StatAssist\\ GradBoost \end{tabular}}  \\
                                  &                            
                                  &
                                  &                                  
                                  &        \\ 
\midrule
%ResNet50~\cite{he2016deep}              & 25.55M& 16.34B & 77.5 & 77.5 & 77.5 & 77.5 \\
ResNet18~\cite{he2016deep}              & 11.68M& 1.82B & 69.7 & 68.8 & 68.9 & 69.6 \\
MobileNetV2~\cite{mobilenetv2}          & 3.51M & 320.2M & 71.8 & 70.3 & 70.7 & 71.5 \\
\midrule
ShuffleNetV2~\cite{Ma_2018_ECCV}        & 2.28M & 150.6M & 69.3 & 63.4 & 67.7 & \textbf{68.8} \\
ShuffleNetV2$\times$0.5~\cite{Ma_2018_ECCV}   & 1.36M & 43.65M & 58.2 & 44.8 & 56.8 & \textbf{57.3} \\
\midrule
\end{tabular}
\caption
{
Classification results~(Top 1 accuracy) on the ImageNet~\cite{russakovsky2015imagenet} dataset. \textbf{FP}: Full-Precision models with floating-point computations. \textbf{QAT}: Quantized models trained or fine-tuned with quantization-aware-training. 
\textbf{Params}: Number of parameters of each model. 
\textbf{FLOPs}: FLOPs measured w.r.t. a $224 \times 224$ input. 
The performance gap between each quantized model fine-tuned with FP pre-trained weights~(\textbf{QAT Fine-tune}) and its FP counterpart~(\textbf{FP Training}) varies according to the model architecture. Our method~(\textbf{StatAssist} \& \textbf{GradBoost}) effectively narrows the gap, especially for ShuffleNetV2~\cite{Ma_2018_ECCV} structures~(\textbf{Row 3 and 4}).
We used the quantized-version of each model, pre-trained FP weights, and training methodology from \texttt{torchvision}~\cite{pytorch}.%\protect\footnotemark
} 
\label{table:quant_classification}
\end{table*}

\begin{table*}[!htb]
\centering
\tabcolsep=0.10cm
\begin{tabular}{@{}lccccccc@{}}
\toprule
\multirow{2}{*}{Model} & 
\multirow{2}{*}{Params} & 
\multirow{2}{*}{FLOPs} & 
\multirow{2}{*}{FP training} &

\multirow{2}{*}{\begin{tabular}[c]{@{}c@{}} QAT\\ Fine-tune \end{tabular}} &
\multirow{2}{*}{\begin{tabular}[c]{@{}c@{}} StatAssist\\ Only\end{tabular}} &
\multirow{2}{*}{\begin{tabular}[c]{@{}c@{}} StatAssist\\ GradBoost \end{tabular}}  \\
                                  &                            
                                  &
                                  &                                  
                                  &        \\ 
\midrule
T-DSOD~\cite{tdsod}    & 2.17M & 2.24B & 71.5 & 71.4 & 71.9 & \textbf{72.0} \\
SSD-mv2~\cite{liu2016ssd}   & 2.95M & 1.60B & 71.0 & 70.8 & 71.1 & \textbf{71.3} \\
%RetinaFace~\cite{deng2019retinaface}-(50)* & ResNet50 & 24 M & 76.5 & 0.965   & 0.956& 0.904 & \\
%RetinaFace-M*\tablefootnote{For RetinaFace-M, we refer the performance of their arxiv publication, which only containing WIDER Hard validation result.} & MNet & 0.21 M & 1.2 & - & - & 0.782 & \\
%EXTD-R (NS)  & Ours  & 3.7 M   & 31.5  & 0.941    & 0.932    & 0.875 & \\
%\textbf{EXTD-R}   & Ours  & \textbf{0.61 M}   & \textbf{31.5}  & \textbf{0.938} & \textbf{0.925} & \textbf{0.871} & \\
%\textbf{EXTD-R-0.25}   & Ours  & \textbf{0.04} M   & \textbf{2.1}  & \textbf{0.854} & \textbf{0.844} & \textbf{0.774} & \\
\midrule
\end{tabular}
\caption
{
Object detection results~(mAP) on PASCAL-VOC 2007 dataset. \textbf{FP}: Full-Precision models with floating-point computations. \textbf{QAT}: Quantized models trained or fine-tuned with quantization-aware-training. 
\textbf{Params}: Number of parameters of each model. 
\textbf{FLOPs}: FLOPs measured w.r.t. a $300 \times 300$ input. 
While quantized models fine-tuned with FP pre-trained weights~(\textbf{QAT Fine-tune}) show an marginal mAP drop compared to their FP counterparts~(\textbf{FP Training}), the models trained from scratch with our proposed method~(\textbf{StatAssist} \& \textbf{GradBoost}) achieve the mAP gain in the \texttt{INT8} quantization setting. 
}

\label{table:quant_detection}
\end{table*}

\subsection{Classification}
\label{sec:classification}
We first compare the \textit{classification} performance of different lightweight models on the ImageNet~\cite{russakovsky2015imagenet} dataset in Table~\ref{table:quant_classification}.We used the quantized-version of each model, pre-trained FP weights, and training methodology from \texttt{torchvision}~\cite{pytorch} 0.6.0.  
We found out that the performance gap between a quantized model fine-tuned with FP weights and each FP counterpart varies according to the architectural difference. In particular, the channel-shuffle mechanism in ShuffleNetV2~\cite{Ma_2018_ECCV} seems to widen the gap. Our method successfully narrows the gap to no more than \textit{$0.9\%$}, proving that the scratch training of \textit{fake-quantized}~\cite{jacob2018quantization} models with StatAssist and GradBoost is essential for better quantized performance.

\subsection{Object Detection}
\label{sec:obj_detection}
For the \textit{object detection}, we used two lightweight-detectors: SSD-Lite-MobileNetV2 (SSD-mv2) \cite{mobilenetv2} and Tiny-DSOD (T-DSOD)~\cite{tdsod}. We trained the models with Nesterov-momentum SGD~\cite{pmlr-v28-sutskever13} on PASCAL-VOC 2007~\cite{pascal-voc-2007} following default settings of the papers. For training T-DSOD, we set the initial learning rate $lr=2e-2$ and scaled the rate into $0.1$ at the iterations 120K and 150K, over entire 180K iteration. In SSD-mv2 training case, we used total 120K iteration with scaling at 80K and 100K. 
The initial learning rate was set to $1e-2$.
For each case, we set the batch size of $64$.
For testing, we slightly modified the detectors to fuse all the layers in each model.

Table~\ref{table:quant_detection} shows the evaluation results on two light-weight detectors, T-DSOD and SSD-mv2. Following our theoretical analysis, the quantized model trained with pre-trained FP weight fine-tuning could not surpass the performance of the FP model, which acts like an upper-bound. On the contrary, we can see that it is possible to make the quantized outperform the original FP by training each model from scratch using our method. 

This is counter-intuitive in that there still exists enough room for improvements in the FP's representational capacity. However, our method still can't be a panacea for any \texttt{INT8} conversion since the model architecture should be modified due to limitations explained in Section~\ref{sec:static_quant}. This modification would induce a performance degradation if the FP model was not initially designed for the quantization.

\begin{table*}[!htb]
\centering
\tabcolsep=0.10cm
\begin{tabular}{@{}lcccccc@{}}
\toprule
\multirow{2}{*}{Model} & 
\multirow{2}{*}{\begin{tabular}[c]{@{}c@{}} Params \end{tabular}} &
\multirow{2}{*}{\begin{tabular}[c]{@{}c@{}} FLOPs \end{tabular}} &
\multirow{2}{*}{FP Training} &
\multirow{2}{*}{\begin{tabular}[c]{@{}c@{}} QAT\\ Fine-tune \end{tabular}} &
\multirow{2}{*}{\begin{tabular}[c]{@{}c@{}} StatAssist\\Only \end{tabular}} &
\multirow{2}{*}{\begin{tabular}[c]{@{}c@{}} StatAssist \\GradBoost \end{tabular}}  \\
                                  &                            
                                  &
                                  &                                  
                                  &        \\ 
\midrule
ESPNet~\cite{mehta2018espnet}    & 0.60M & 30.2B & 65.4  & 64.6 & 65.0 & 65.5 \\
ESPNetV2~\cite{mehta2018espnetv2} & 3.43M & 48.6B  & 64.4  & 63.8 & 64.6 & 64.5 \\
Mv3-LRASPP-Large~\cite{howard2019searching}  & 2.42M & 12.8B  & 65.3 & 64.5 & 64.7 & 65.2 \\
Mv3-LRASPP-Small~\cite{howard2019searching}    & 0.75M & 3.95M & 62.5 & 61.7 & 61.6 & 62.1 \\
\midrule
Mv3-LRASPP-Large-RE (Ours)   & 2.42M & 12.8B & 65.5  & 64.9 & 65.1 & \textbf{65.8} \\
Mv3-LRASPP-Small-RE (Ours)   & 0.75M & 3.95M & 61.5  & 61.2 & 62.1 & \textbf{62.3} \\

\midrule
\end{tabular}
\caption{Semantic segmentation (mIOU) on Cityscapes \textit{val} set. \textbf{FP}: Full-Precision models with floating-point computations. \textbf{QAT}: Quantized models trained or fine-tuned with QAT.  \textbf{RE}: Replace the \textit{hard-swish} activation~\cite{howard2019searching} with \textit{ReLU}~\cite{pmlr-v15-glorot11a}. \textbf{Large \& Small}: Different model configurations targeted at high and low resource use cases respectively. \textbf{Params}: Number of parameters of each model. \textbf{FLOPs}: FLOPs measured w.r.t. a $2048 \times 1024$ input. While quantized models fine-tuned with FP pre-trained weights~(\textbf{QAT Fine-tune}) show an average \textit{0.65\% mIOU} drop compared to their FP counterparts~(\textbf{FP Training}), models trained from scratch with our proposed method~(\textbf{StatAssist} \& \textbf{GradBoost}) achieve an average \textit{0.13\% mIOU gain} even in the \texttt{INT8} quantization setting. Row 5 and 6 are promising segmentation candidates for an edge-device with high and low resource capacity accordingly.}
\label{table:quant_segmentation}
\end{table*}

\begin{figure*}[!htb]
  \centering
\begin{subfigure}{0.45\textwidth}
  \centering
\includegraphics[width=\linewidth]{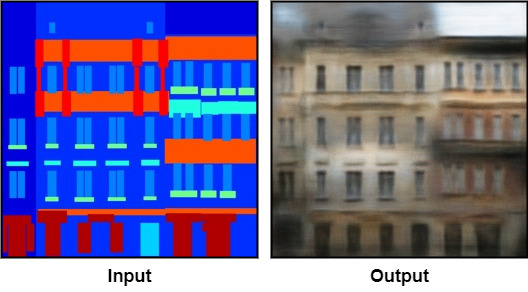}
\caption{Labels to Facade}
\label{fig:3_1}
\end{subfigure}
\begin{subfigure}{0.45\textwidth}
  \centering
\includegraphics[width=\linewidth]{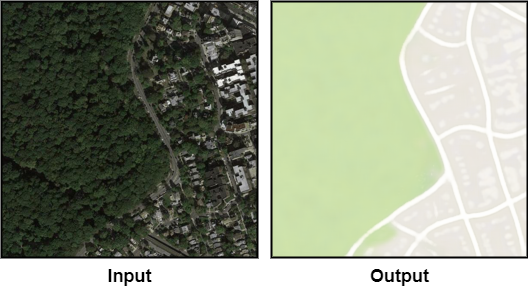}
\caption{Aerial to Map}
\label{fig:3_2}
\end{subfigure}
\begin{subfigure}{0.45\textwidth}
  \centering
\includegraphics[width=\linewidth]{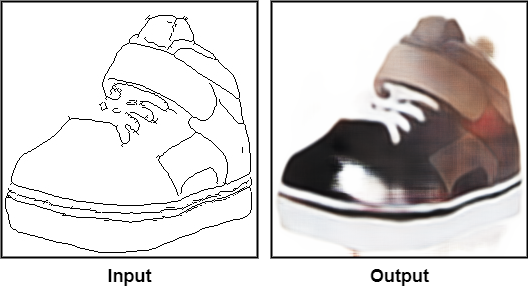}
\caption{Edges to Shoes}
\label{fig:3_3}
\end{subfigure}
\begin{subfigure}{0.45\textwidth}
  \centering
\includegraphics[width=\linewidth]{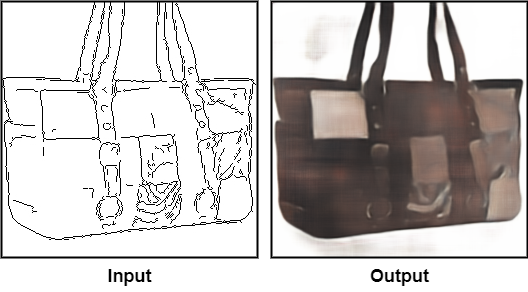}
\caption{Edges to Handbag}
\label{fig:3_4}
\end{subfigure}
\vspace{-2mm}
\caption{Examples results of the quantized Pix2Pix~\cite{pix2pix} model on several image-to-image style transfer problems. The proposed StatAssist and GradBoost enables the scratch training of Pix2Pix model with minimax image generation loss even in the \textit{fake-quantized}~\cite{fan2020training} condition.}
\label{fig:3}
\end{figure*}
\begin{figure*}[!htb]
  \centering
\begin{subfigure}{0.24\textwidth}
  \centering
\includegraphics[width=\linewidth]{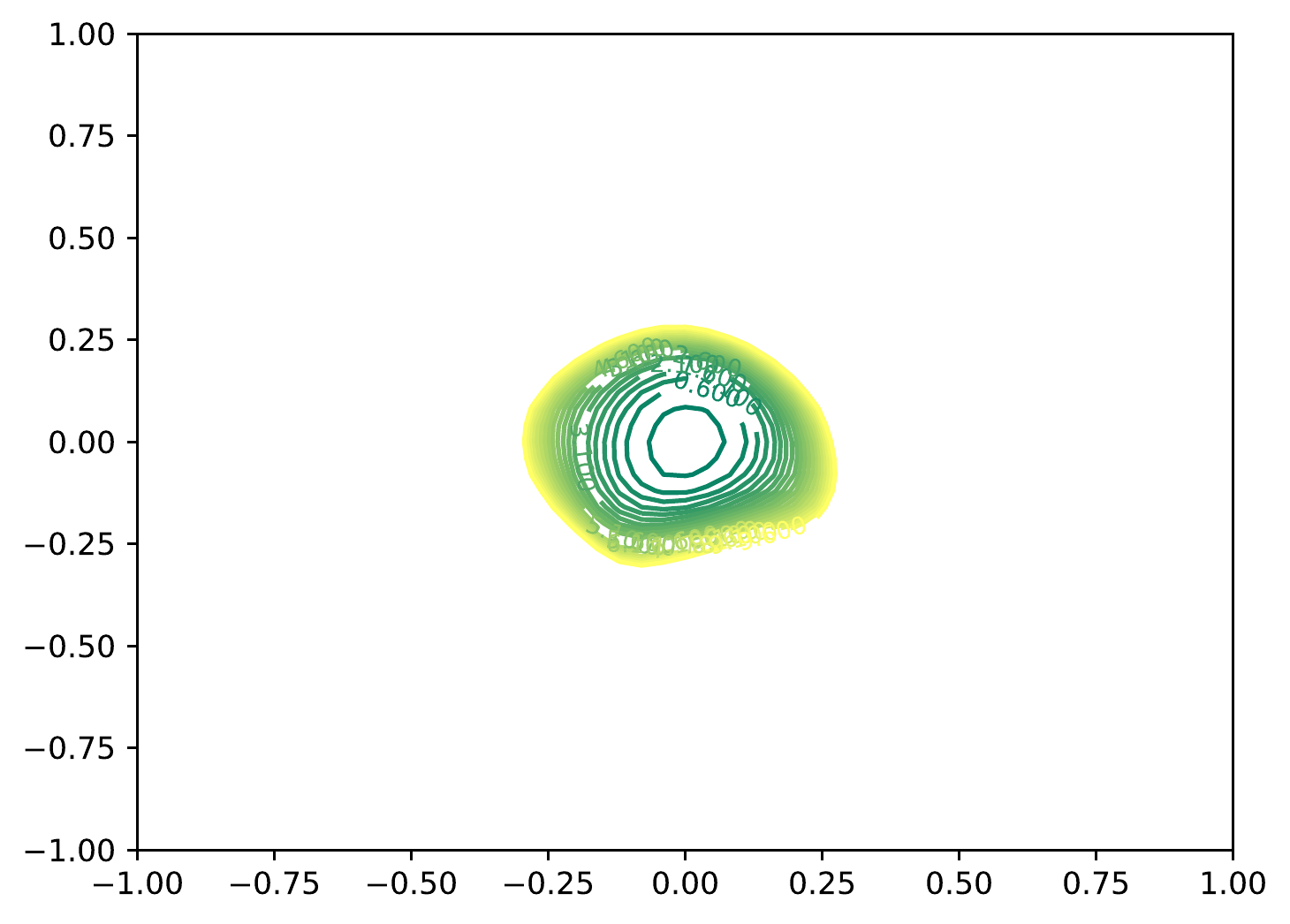}
\caption{FP Training}
\label{fig:2_1}
\end{subfigure}
\begin{subfigure}{0.24\textwidth}
  \centering
\includegraphics[width=\linewidth]{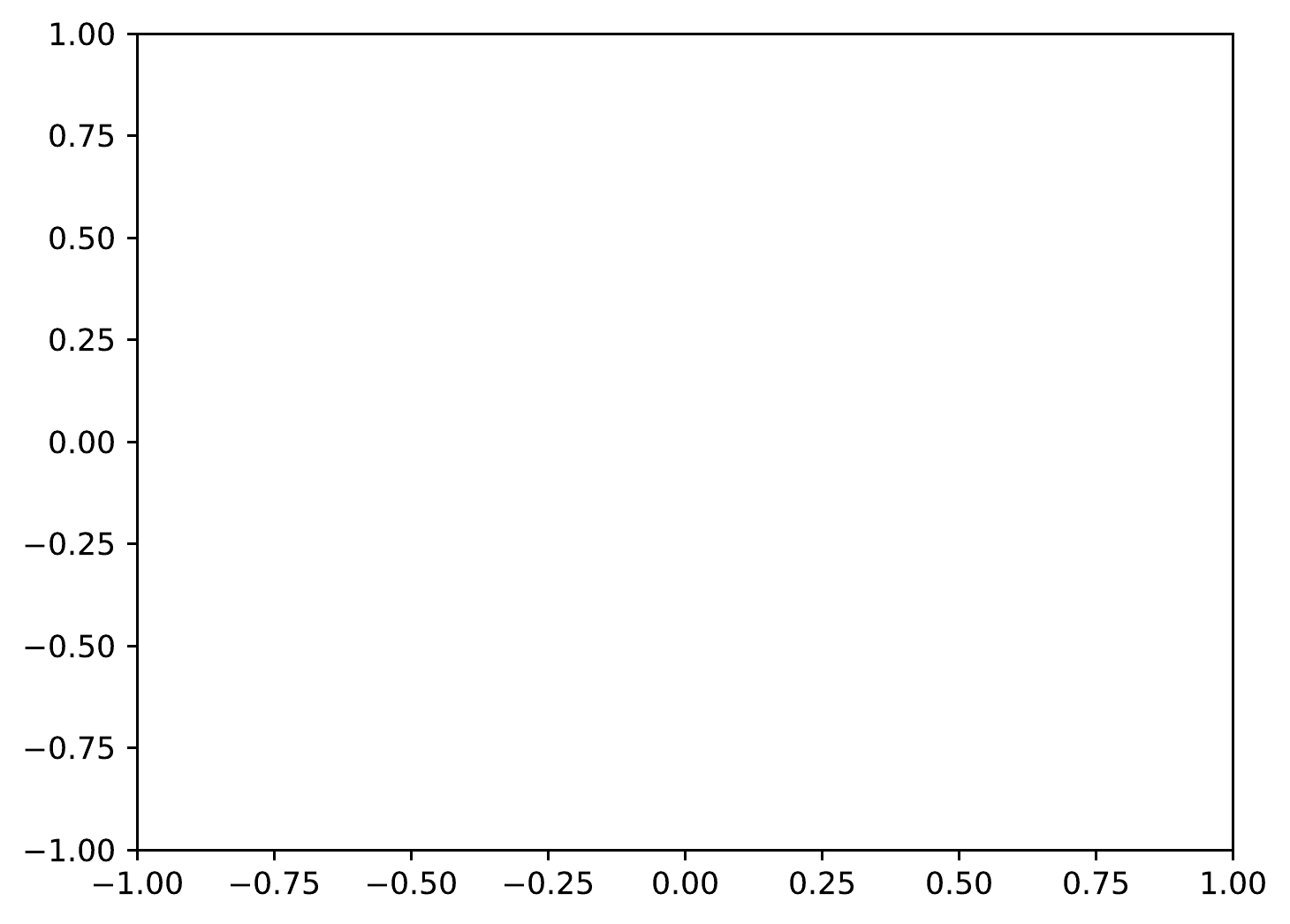}
\caption{Scratch QAT}
\label{fig:2_2}
\end{subfigure}
\begin{subfigure}{0.24\textwidth}
  \centering
\includegraphics[width=\linewidth]{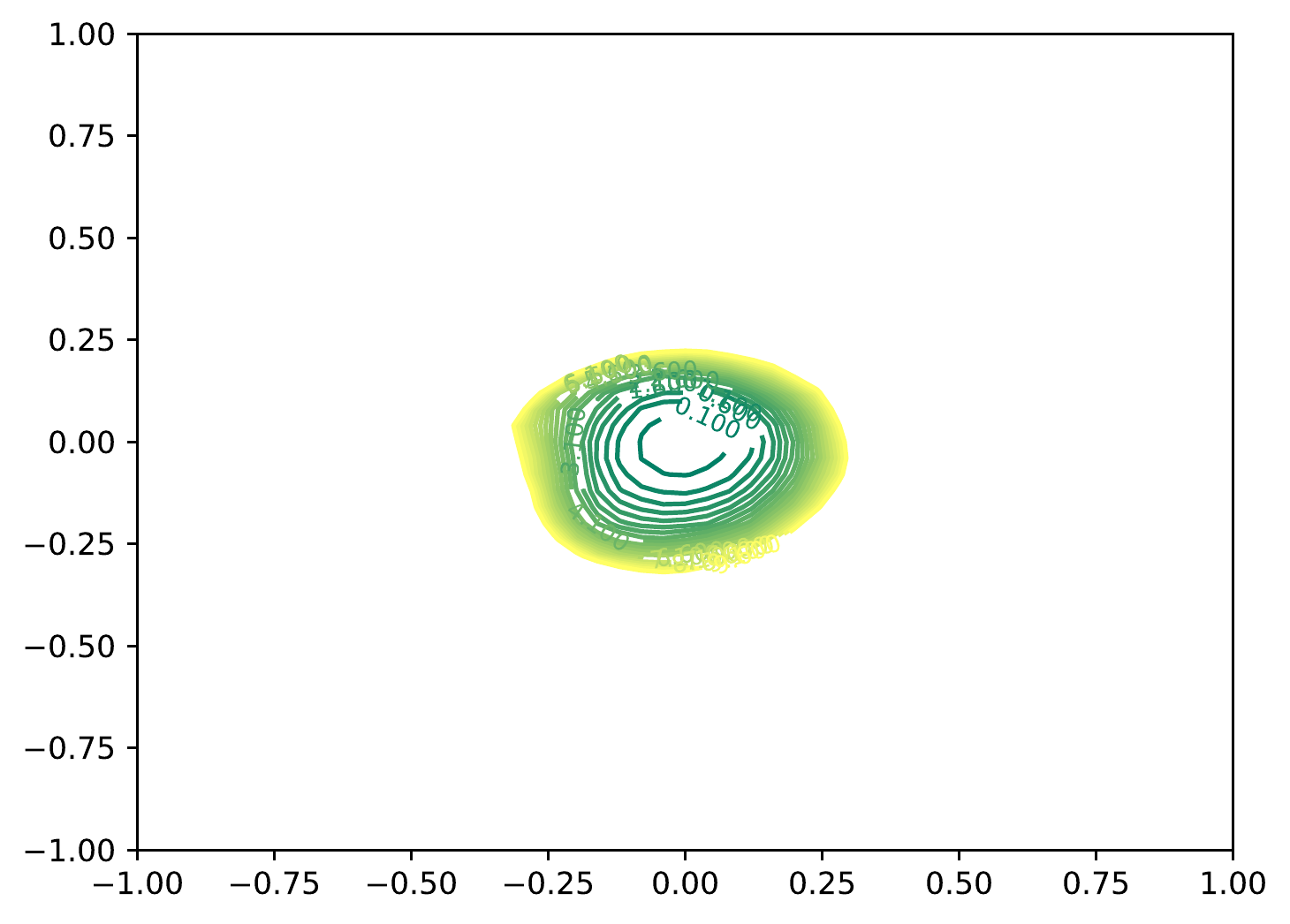}
\caption{StatAssist-only}
\label{fig:2_3}
\end{subfigure}
\begin{subfigure}{0.24\textwidth}
  \centering
\includegraphics[width=\linewidth]{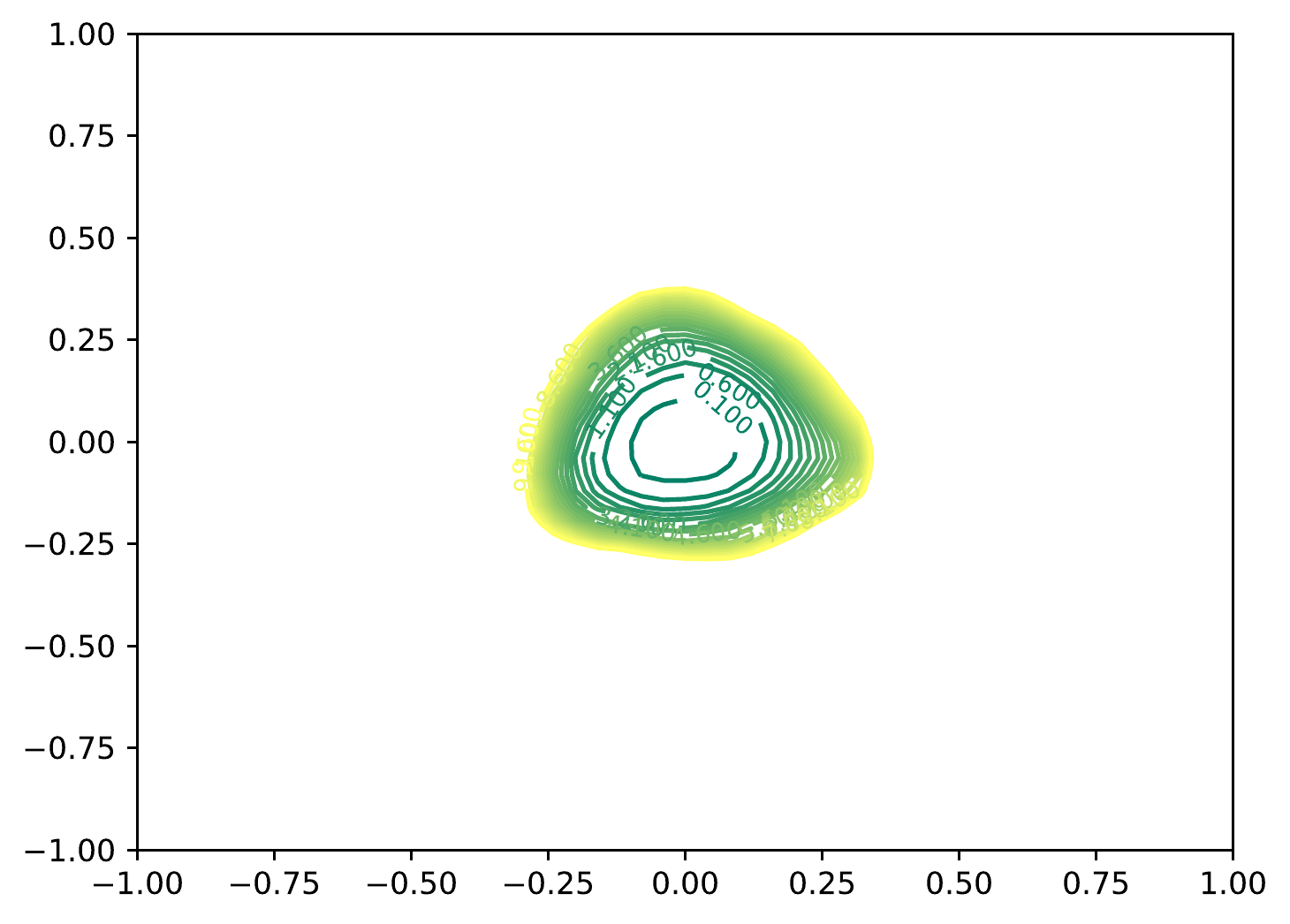}
\caption{StatAssist + GradBoost}
\label{fig:2_4}
\end{subfigure}
\caption{Loss landscapes~\cite{loss_landscape} of a MobileNetV2~\cite{mobilenetv2} on the CIFAR-10~\cite{cifar} with full-precision~(FP) training, original scratch quantization-aware training~(QAT)~\cite{jacob2018quantization},  StatAssist-only, and StatAssist + GradBoost QAT~(\textit{left to right}). Each StatAssist-only and StatAssist + GradBoost QAT model shows a stable loss landscape similar to that of the FP-trained model while the scratch QAT model draws a flat surface. The StatAssist + GradBoost QAT model also covers a wider range of the loss terrain, indicating that the GradBoost method broadens the search area for optimal local-minima.}
\label{fig:2}
\end{figure*}
\subsection{Semantic Segmentation}
\label{sec:exp_semantic_seg}
We also evaluated our method on \textit{semantic segmentation} with three lightweight-segmentation models: ESPNet~\cite{mehta2018espnet}, ESPNetV2~\cite{mehta2018espnetv2}, and MobileNetV3 + LRASPP~(Mv3-LRASPP)~\cite{howard2019searching}. We trained the models on Cityscapes~\cite{Cordts2016Cityscapes} following default settings from~\cite{DBLP:journals/corr/abs-1802-02611}. For training, we used Nesterov-momentum SGD~\cite{pmlr-v28-sutskever13} with the initial learning rate $lr=7e-3$ and \textit{poly} learning rate schedule~\cite{howard2019searching}. We trained our models with $768 \times 768$ random-cropped \textit{train} images to fit a model in a single NVIDIA P40 GPU. The evaluation was performed with full-scale $2048 \times 1024$ \textit{val} images. For Mv3-LRASPP, we also made extra variations to the original architecture settings from ~\cite{howard2019searching} (as in our supplementary material) to examine promising performance-compression trade-offs.

The segmentation results in Table~\ref{table:quant_segmentation} are in consensus with the results in \ref{sec:obj_detection}. While quantized models fine-tuned with FP weights suffer from an average \textit{$0.65\%$ mIOU drop} compared to their FP counterparts, the StatAssist + GradBoost trained models maintain or slightly surpass the performance of the FP with an average \textit{$0.13\%$ mIOU gain}. While it is cost-efficient to use \textit{hard-swish} activation~\cite{howard2019searching} in the FP versions of the MobileNetV3,the \textit{Add} and \textit{Multiply} operations used in \textit{hard-swish} seems to generate extra quantization errors during the training and degrade the final quantized performance. Our modified version of MobileNetV3 (Mv3-LRASPP-Large-RE, Mv3-LRASPP-Small-RE), in which all \textit{hard-swish} activations are replaced with the ReLU, states that the right choice of activation function is important for the quantization-aware model architecture.

\subsection{Style Transfer}
\label{sec:style_transfer}

We further evaluate the robustness of our method against unstable training losses by training the \textit{Pix2Pix}~\cite{pix2pix} \textit{style transfer} model with minimax~\cite{NIPS2014_5423} generation loss. For the \textit{layer fusion} compatibility, we used ResNet-based Pix2Pix model proposed by Li \etal~\cite{li2020gan} and Adam \cite{kingma2014adam} optimizer with our StatAssist and GradBoost. We only applied the \textit{fake-quantization}~\cite{fan2020training} to the model's \textit{Generator} since the \textit{Discriminator} is not used during the inference. Example results on several image-to-image style transfer problems are in Figure~\ref{fig:3}. We demonstrate that our method also fits well to the \textit{fuzzy} training condition without causing the \textit{mode collapse}~\cite{NIPS2014_5423}, which is considered as a sign of failure in minimax-based generative models. As demonstrated in Figure ~\ref{fig:3}, our method successfully trained the Pix2Pix model on different image-to-image style transfer problems.

\section{Possible Considerations for Quantization-Aware Model Designing and Training}

From the above results, we can raise an issue regarding the importance of the full-precision~(\texttt{FLOAT32}) pre-trained model to initiate quantization-aware training. 
Previous works have assumed that the loss surface of a quantized (\texttt{INT8}) model is the approximated version of the loss surface of full-precision~(\texttt{FLOAT32}) model, and hence, been focusing on fine-tuning of \textit{fake-quantized} model to narrow the approximation gap.
Our observations, however, show a new possibility that the quantized loss surface itself has a different and better local minima.
In above experiments, we show that using only a single epoch in full-precision~(\texttt{FLOAT32}) setting to warm-up the optimizer with a proper direction of gradient momentum can achieve comparable or better results than using the full-precision~(\texttt{FLOAT32}) pre-trained weight. As shown in Figure~\ref{fig:2_4}, the combination of StatAssist and GradBoost stabilizes the training and broadens the search area for optimal local minima during QAT.

\end{document}